\documentclass[10pt,twocolumn,letterpaper]{article}

\usepackage[pagenumbers]{cvpr} 

\usepackage{algorithm}
\usepackage{listings}
\usepackage[dvipsnames]{xcolor}
\usepackage{algpseudocode}

\usepackage{colortbl}
\definecolor{Gray}{gray}{0.90}

\definecolor{codeblue}{rgb}{0.25, 0.5, 0.5}
\definecolor{codeorange}{rgb}{1.0, 0.5, 0.3}
\definecolor{codegreen}{rgb}{0.13, 0.54, 0.13}

\definecolor{cvprblue}{rgb}{0.21,0.49,0.74}
\usepackage[pagebackref,breaklinks,colorlinks,citecolor=cvprblue]{hyperref}

\usepackage{graphicx}
\usepackage{makecell}
\usepackage{hhline}
\usepackage{comment}

\def\paperID{} 
\def\confName{CVPR}
\def\confYear{}

\title{Diffusion Illusions: Hiding Images in Plain Sight}

\author{%
\vspace{-0.5em}
  Ryan Burgert \quad
  Xiang Li \quad 
  Abe Leite \quad
  Kanchana Ranasinghe \quad
  Michael S. Ryoo
  \vspace{0.5em} \\
  Stony Brook University \quad 
  \vspace{0.2em} \\
  \small{\texttt{rburgert@cs.stonybrook.edu}}
  \vspace{0.8em} 
}

\newcommand{\projurl}{\url{https://diffusionillusions.com}}

\begin{document}

\twocolumn[{%
\renewcommand\twocolumn[1][]{#1}%
    \vspace{-5.8em}
\begin{center}
\maketitle
    \vspace{-2.8em}
    \centering
    \includegraphics[width=1\textwidth]{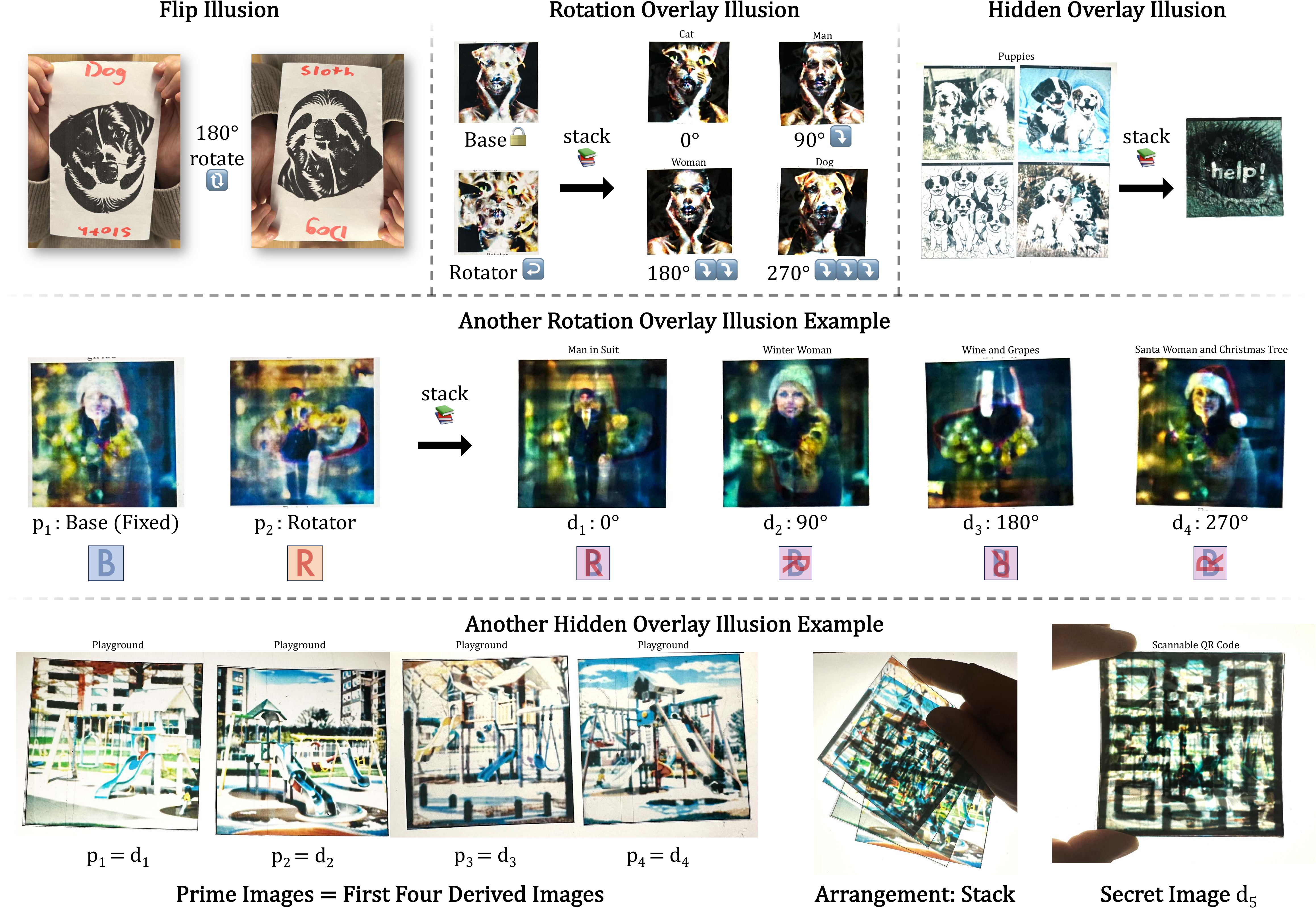}
    \vspace{-1.3em}
    \captionof{figure}{Diffusion Illusions are a new class of automatically generated optical illusions. The images on top demonstrate the three major types of illusions we discuss in this paper: Flip Illusions, Rotation Overlay Illusions, and Hidden Overlay Illusions. (Terminology is formally defined in \cref{sec:problem}). The bottom showcases an example of Hidden Overlay Illusions: four images (prime images $p_{1\dots4}$) that when stacked on top of each other (arrangement) reveal a new fifth image (derived image $d_5$). 
    \textit{Please note that these illustrations are \underline{all photographs} of the generated images \underline{physically fabricated} in the real world.} 
    }
    \label{fig:teaser}
    \vspace{0.5em}
\end{center}%
}]

\begin{abstract}
\vspace{-1.4em}
We explore the problem of computationally generating special `prime' images that produce optical illusions when physically arranged and viewed in a certain way.
First, we propose a formal definition for this problem.
Next, we introduce Diffusion Illusions, the first comprehensive pipeline designed to automatically generate a wide range of these illusions.
Specifically, we both adapt the existing `score distillation loss' and propose a new `dream target loss' to optimize a group of differentially parametrized prime images, using a frozen text-to-image diffusion model.
We study three types of illusions, each where the prime images are arranged in different ways and optimized using the aforementioned losses such that images derived from them align with user-chosen text prompts or images.
We conduct comprehensive experiments on these illusions and verify the effectiveness of our proposed method qualitatively and quantitatively.
Additionally, we showcase the successful physical fabrication of our illusions --- as they are all designed to work in the real world.
Our code and examples are publicly available at our 
\textit{interactive} project website: \projurl
\vspace{-1.5em}
\end{abstract}

\section{Introduction}
An image that is viewed right-side up appears to be an ordinary photo of a dog but viewed upside-down looks like a sloth. Four images, each showing an everyday playground, when superimposed form a QR code (see \cref{fig:teaser}). These types of images that cause illusions have long required immense time and skill to create, but we have developed a general pipeline capable of generating appealing illusions automatically.
More specifically, given a frozen text-to-image diffusion model, we adapt existing score distillation loss and propose a new dream target loss to optimize a group of prime images differentiably parametrized by fourier feature networks.
Eventually, the images are optimized to comply with the textual and/or image prompts given by the user to trigger illusions in a certain arrangement.

Generating such images is not the sole domain of play. Illusions -- that is, visual stimuli whose interpretation depends on how they are arranged and viewed -- have been created and studied for centuries. While they are an appealing sort of ``visual puzzle'', they also reveal much about how humans perceive the world and about the abstract structure of images. Even though illusions have been created and studied for centuries, and certain types have been generated by computers for decades, photorealistic illusions have remained largely out of reach until the very recent past, and until this point, there has been no general framework for understanding and generating such illusions.

\subsection{Contributions}
In this paper, we present the first formalized, generic framework for creating such illusions. We name our framework \textit{Diffusion Illusions}. Our major contributions can be summarized as follows:
\begin{enumerate}[leftmargin=1.5em,noitemsep,topsep=0.0ex,itemsep=-1.0ex,partopsep=0ex,parsep=1ex]
   \item We provide the first formal definition for the problem of generating illusions; 
   \item We present Diffusion Illusions, a flexible tool for generating multiple types of illusions;
   \item We assess the quality of computer-generated illusions in multiple aspects and conduct comprehensive experiments to validate the effectiveness of our method;
   \item We successfully fabricate the generated images and their corresponding illusions in the real world.
\end{enumerate}

\subsection{Related Work: History of Illusions}
\begin{figure}
\centering
\includegraphics[width=1\linewidth]{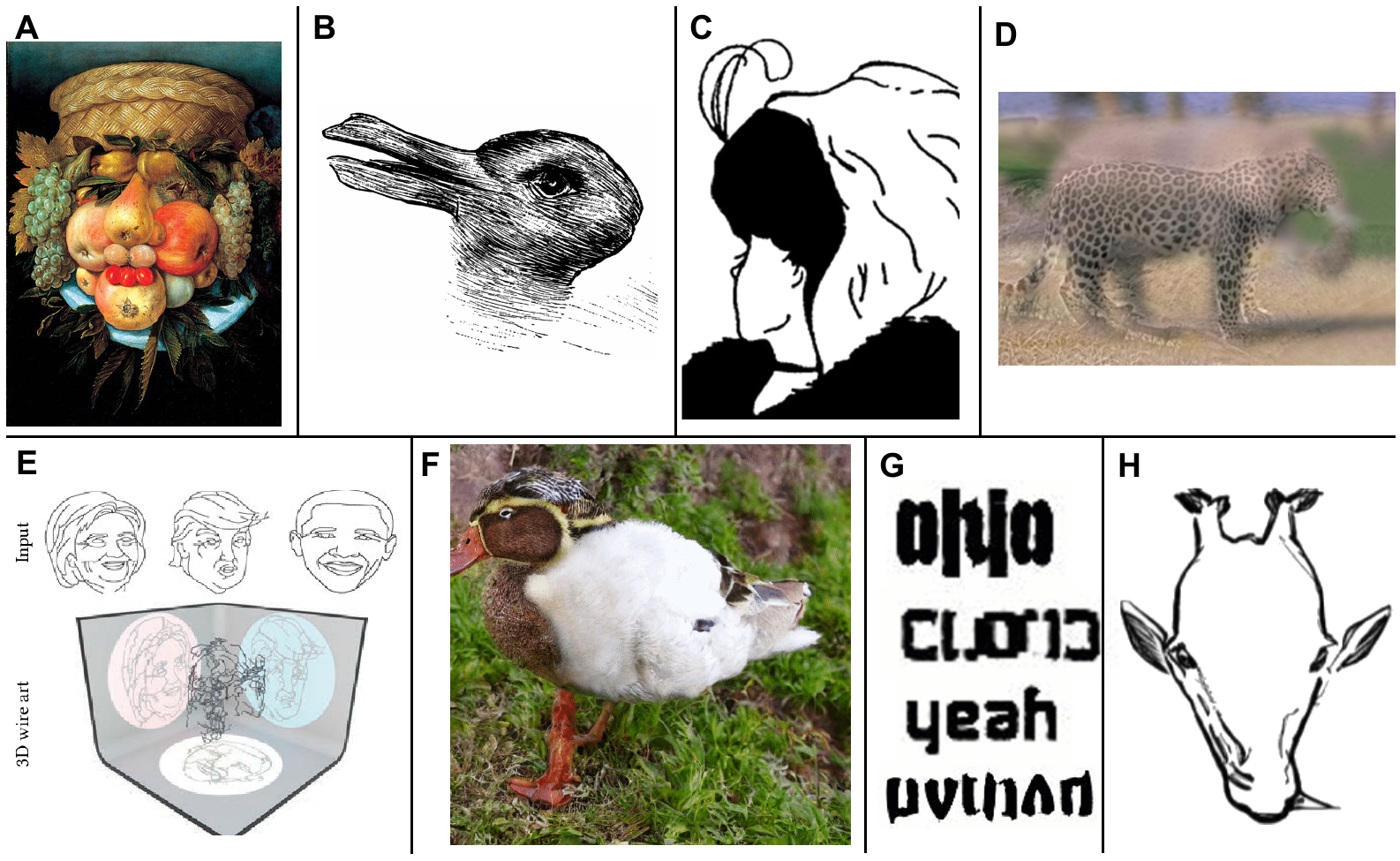}
\vspace{-2.0em}
\caption{\small
A brief history of illusions. \textbf{Classical illusions}: (A) ``Fruit Basket'' (1500s) by Giuseppe Arcimboldo provides a very early example, depicting a face when viewed in one orientation and a fruit basket when viewed in the other. (B) When viewed directly, ``Kaninchen und Ente'' (1892) is ambiguous; $45^\circ$ rotations make it appear as a rabbit or a duck \cite{jastrow1899mind, wittgenstein_philosophical_1953}. (C) ``My Wife and My Mother-in-Law'' (1915) by William Ely Hill may be interpreted as showing either a young or an old woman depending on how it is grouped \cite{boring_new_1930, nicholls_perception_2018}. \textbf{Computationally-generated illusions}: (D) a hybrid image which appears to be a leopard when viewed close-up and an elephant when viewed from a distance \cite{oliva2006hybrid}. (E) a wire sculpture which depicts three different 2010s American politicians when viewed from different angles \cite{hsiao_multi-view_2018}. \textbf{Diffusion-based illusions}: (F) an image depicting a duck when viewed upright and a rabbit when rotated $90^\circ$ ccw \cite{tancik_illusiondiffusion_2023}. (G) a set of computationally-generated ambigrams reading `Ohio', `cloud', `yeah', and `python' \cite{samsudin_ambigrams_2023}. (H) an image depicting a giraffe when viewed upright and a penguin when viewed upside-down \cite{geng2023visualanagrams}.}
\label{fig:intro}
\vspace{-1em}
\end{figure}

\subsubsection{Classical illusions}
Images whose interpretation depends on viewing angle or category bias, sometimes known as ambiguous images, have been designed for centuries. Such images have drawn the scholarly interest of psychologists \cite{jastrow1899mind, boring_new_1930} and philosophers \cite{wittgenstein_philosophical_1953} since the 1800s. Ambiguous images have been used experimentally to understand how category bias during perception varies as people age \cite{nicholls_perception_2018}, and families of ambiguous images, such as ambigrams \cite{hofstadter1985meta}, are often constructed as a way of better understanding the domains they belong to. We present some relevant examples of classical illusions in \Cref{fig:intro}.

\subsubsection{Computationally-generated illusions}
A growing stream of research has focused on computationally generating specific types of illusions. One early example is hybrid images \cite{oliva2006hybrid}. Hybrid images are created from two images by combining the low-frequency features of one with the high-frequency features of the other. Viewers see the object from the low-frequency image when viewing the hybrid image from a distance, and see the object from the high-frequency image when viewing up-close. While this process may be automated, the authors note that for best results, the overall shapes of the low-frequency and high-frequency images should be manually aligned.

A number of researchers have created 3-dimensional objects that are interpreted as different objects when they are viewed from different angles. In multi-view wire art \cite{hsiao_multi-view_2018}, a single 3D wire may be viewed or lit from multiple angles to obtain different clean line drawings; and in view-dependent surfaces \cite{perroni-scharf_constructing_2023}, a colored 3D-printed height field may be viewed from different angles to obtain different colored images.

An additional type of illusion is steganography, in which apparently normal objects may be viewed in a particular way to uncover a hidden meaning. In The Magic Lens \cite{papas_magic_2012}, seemingly meaningless dots are generated such that, when viewed through an intricate refractive lens, they will comprise a specified image.

\subsubsection{Diffusion-based Image Generation}
Diffusion Probabilistic Models \cite{pmlr-v37-sohl-dickstein15} resulted in rapid advances for image generation tasks, including text-to-image generation \cite{Nichol2022GLIDETP,diffusionbeatsgans,dalle,dalle2,imagen,palette,parti,sr3}. Recent works \cite{Poole2022DreamFusionTU,Burgert2022PeekabooTT} sample pre-trained diffusion models without re-training to generate outputs in novel domains. Score Distillation introduced in DreamFusion \cite{Poole2022DreamFusionTU} is the underlying technique enabling optimization of samples in any arbitrary parameter space without backpropagation through the diffusion model. 
We utilize these techniques to construct a novel framework for illusion generation. 
These rapid advances have led to an exploration of suitable evaluation metrics, both quantitative and qualitative \cite{lee_holistic_2023,Benny2020EvaluationMF,Betzalel2022ASO,yeh2023navigating, friedman2022vendi}, which we use to evaluate our proposed framework.

\subsubsection{Contemporary Work}
Following recent image generation developments, a small but growing body of non-scholarly or unpublished work has approached the problem of generating multi-view 2D images \cite{tancik_illusiondiffusion_2023} or ambigrams \cite{samsudin_ambigrams_2023}. While these approaches appear to yield appealing results, they are narrowly focused on specific illusions, may require substantial cherry-picking, and have not been formally presented or published. 
In contrast, we present a formalized, generic approach capable of generating variable types of illusions followed by extensive evaluation (both quantitative and qualitative) of our approach.
Inspired by our Diffusion Illusions project, contemporary work in \cite{geng2023visualanagrams} presents a formal framework for efficient (fast inference) illusion generation, but operates on a subset of our illusions (namely, those with a single ``prime image'' in our terminology). Furthermore, \cite{geng2023visualanagrams} does not explore any illusions with overlay which is generally more challenging and the generality for real-world transfer (i.e. fabrication of illusions in the real world). 


\section{Problem Statement}
\label{sec:problem}

\begin{figure*}
\centering
\includegraphics[width=0.7\linewidth]{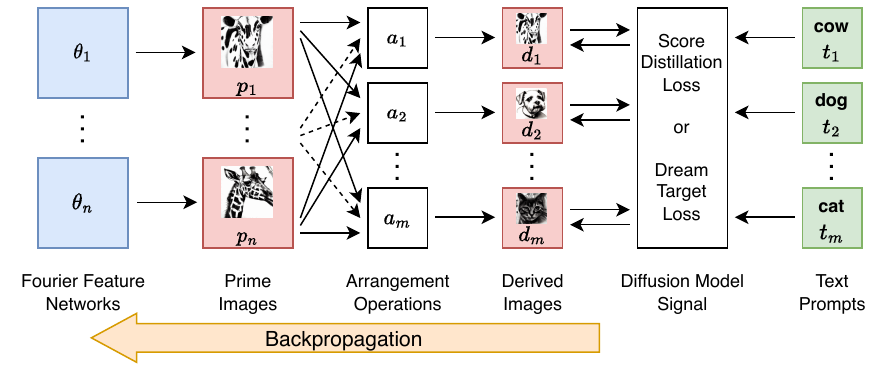}
\vspace{-1.0em}
\caption{Architecture overview. Trainable components are shown in blue, intermediate variables are in red, non-trainable functions are in white, and inputs are in green. A diffusion network provides two different loss signals pulling the derived images towards the text prompts. Only a single loss signal, either Score Distillation Loss or Dream Target Loss, is computed at each training step. Gradients on the derived images are backpropagated through the arrangement operations and prime images to the parameters of the Fourier Feature Networks. No backpropagation occurs through the diffusion network.}
\label{fig:overview}
\vspace{-1em}
\end{figure*}
We define an illusion as the situation that occurs when a set of physical images called \textit{prime images} $p$ are viewed or \textit{arranged} in multiple ways, with each arrangement yielding a unique perceived image, referred to as a \textit{derived image} $d$, that represents a specific object or scene. 

Most of the existing illusions we have discussed consist of a single 2D image or 3D object as a prime image, with the arrangements being simple translations and rotations of the prime image in 2D or 3D space. In the simplest case where a 2D drawing is rotated to yield different perceived objects, the arrangement operations may be modeled as simple rotations. The near and distant views composing the Hybrid Images illusion \cite{oliva2006hybrid}, on the other hand, might be best modeled by high-pass and low-pass spatial frequency filters.

In an effort to find a fully general definition of illusions and leverage the new possibilities afforded by text-to-image models, we do not limit ourselves to a single prime image. 
We additionally consider situations where \textit{multiple} composable prime images, for instance, stencils or light-filtering transparencies, may be arranged in different ways to yield different derived images. In the particular case of composing two light-filtering transparencies, the arrangement operation may be modeled as a rotation of each prime image followed by a multiply operation to model the light-filtering step.

Formally, the illusion process is described as follows. Consider some prime image space $\mathcal{P}$ representing physically realizable visual stimuli, and some derived image space $\mathcal{D}$ representing a human view of a scene. (Practically, we use 2D RGB images to represent both spaces.) Then, an illusion consists of a tuple of $n$ prime images $\{p_1, p_2, \ldots, p_n\}, p_i \in \mathcal{P}$ and a tuple of $m$ arrangement operations $A = \{a_1, a_2, \ldots, a_m\}, a_j : \mathcal{P}^n \rightarrow \mathcal{D}$. Each $a_j$ represents an arrangement of all of the prime images to obtain a single derived image $d_j$, such that the illusion yields a tuple of $m$ derived images $\{d_1, d_2, \ldots, d_m\}, d_j \in \mathcal{D}$. (This articulation may be easily generalized to heterogeneous illusions, such as a wireframe viewed through a stencil; in this case, each prime image $p_i$ belongs to its own prime image space $\mathcal{P}_i$.)

This framing is complementary to the existing literature on ``ambiguous images''. The illusion process is not intended to cover images that have multiple interpretations when viewed in exactly the same way, though it may be possible to articulate a perceptual bias towards a certain category as a type of arrangement. However, the illusion process otherwise broadens the category of ambiguous images to include situations involving multiple composed images. We propose multiple examples below that are to our knowledge wholly novel.

This definition allows one to separate the process of creating an illusion into two steps: first, selecting a prime image domain and defining and modeling the arrangement operation; and second, searching the prime image domain for images that yield the desired derived images when arranged in each way. While the first step requires creativity and experimentation, the second is sufficiently concrete that it may be practically automated, as discussed in \cref{sec:method}.


\section{Method}
\label{sec:method}

We introduce Diffusion Illusions, a flexible tool for generating multiple types of visual illusions that can be styled with unprecedented control (e.g. photorealistic images, artistic styles, or even arbitrary information such as QR codes).
At a high level, the Diffusion Illusions pipeline consists of 
\begin{itemize}[leftmargin=2em,noitemsep,topsep=0.0ex,itemsep=-1.0ex,partopsep=0ex,parsep=1ex]
    \item a set of prime images parameterized by $\theta$ ($\mathcal{P}$),
    \item a set of specific arrangement processes ($A$, that derive images from all primes), 
    \item a frozen text-to-image diffusion model ($\mathcal{F}$)
\end{itemize}
We refer to the outputs of the arrangement processes as derived images ($D$). 
The diffusion model is used to provide a signal using one of two mechanisms (\textit{Score Distillation Loss} or \textit{Dream Target Loss}, which will be covered in \cref{sec:losses}) to suitably optimize the prime images, which in turn modifies the derived images.
Our overall pipeline is illustrated in \cref{fig:overview}.

\subsection{Prime Images}
As described in \cref{sec:problem}, prime images are the physical images we eventually want to generate, that will trigger an illusion when viewed or arranged in multiple ways.

In our framework, prime images are represented as $512 \times 512$ dimensional RGB images, meaning that $\mathcal{P} \simeq \mathbb{R}^{(512,512,3)}$.
Instead of direct pixel-space image representation, we use Fourier Features Networks (FFN)~\cite{tancik2020fourier} to represent prime images in parametric form. 
For each prime image, the learnable weights of a single MLP network act as its representation. 
The MLP network maps image-space coordinates to corresponding RGB values similar to \cite{Burgert2022}, forming an implicit image representation.
We further discuss the advantages of FFN in \cref{sec:discussion}.


\subsection{Arrangement Processes}
\label{sec:illusions}
The purpose of arrangement processes, $A$, is to operate on a set of prime images (including single element sets) and produce unique outputs, the derived images. For a single arrangement process $a_i$, 
\begin{align}
    d_i = a_i(P)
\end{align}
each unique sequence of prime images produces a distinct derived image, $d_i$.
Each operation $a_i \in A$ should possess three properties: 
1) For the same set of inputs the operation should always provide the same output (fixed operation). 
2) $a_i$ should also be differentiable, i.e., the possibility to explicitly calculate gradients propagation from output to input through the operation. 
3) $a_i$ should also be realizable in the real world: some series of physical actions on prime images (in physical form) should result in the same derived image. 
To summarize, an arrangement process must be fixed, differentiable, and realizable in the real world. 

\begin{figure}
    \centering
    \includegraphics[width=1\linewidth]{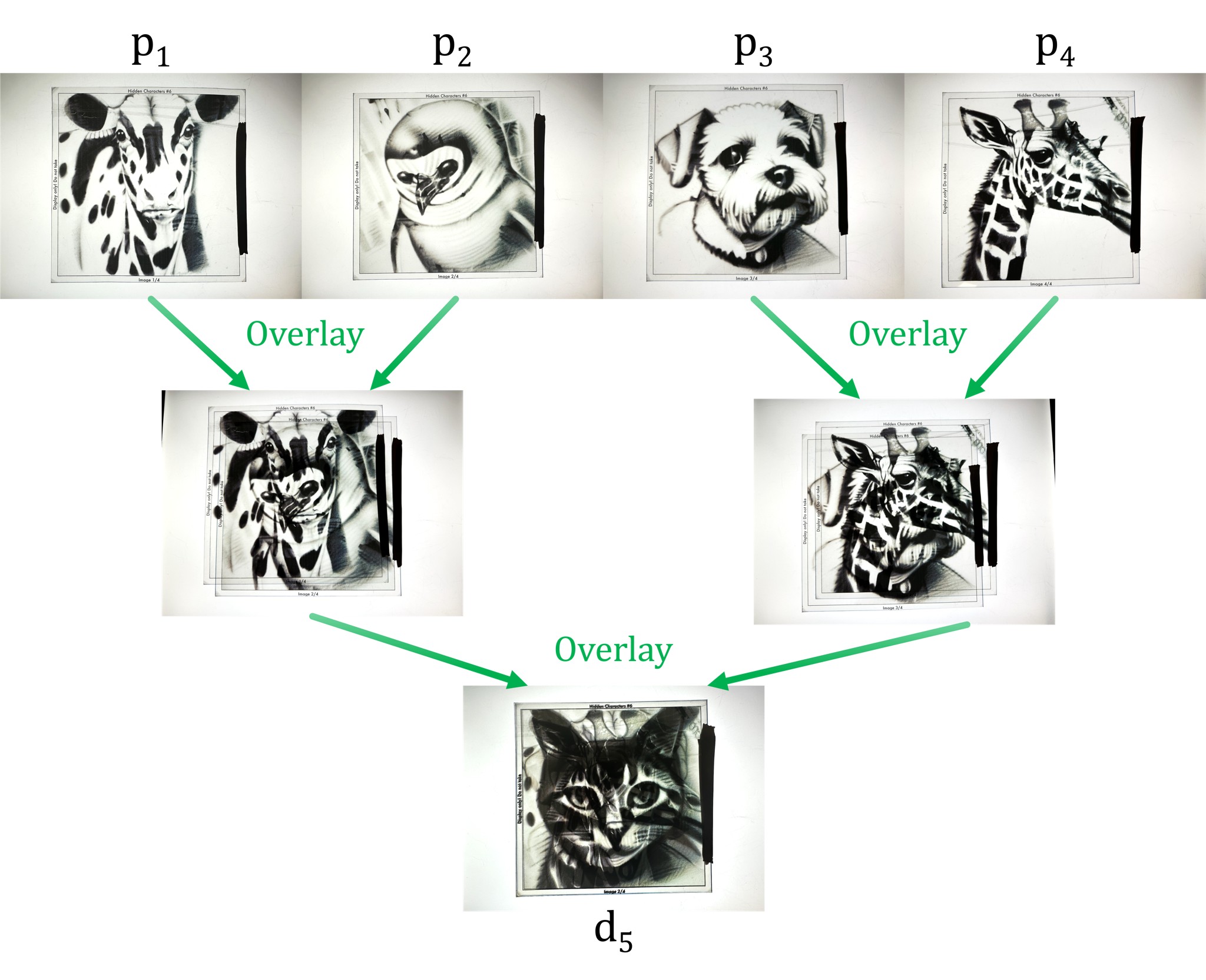}
    \caption{This figure shows the rotation overlay illusion arrangement process. \textit{Please note that these are all \underline{real photographs.}}}
    \label{fig:hidden_overlay_animals_photo}
\end{figure}

We select three illusion categories for further study:

\begin{itemize}

\item \textbf{Flip Illusion} is one of the most classical types of illusions. We define this illusion as consisting of a single 2D prime image, which is interpreted as some object when viewed upright (the first derived image $d_1$) and as another object when viewed upside-down (the second derived image $d_2$).

\item \textbf{Rotation Overlay Illusion} is a minimal type of illusion involving multiple prime images. This illusion is based on two square light-filtering 2D prime images, one base and one rotator. The rotator image is rotated with 0, 90, 180, and 270 degree angles and superimposed on the base image; each rotation yields a derived image interpreted as a different object (see \cref{fig:rot_illusion}).

\begin{figure}
    \centering
    \includegraphics[width=1\linewidth]{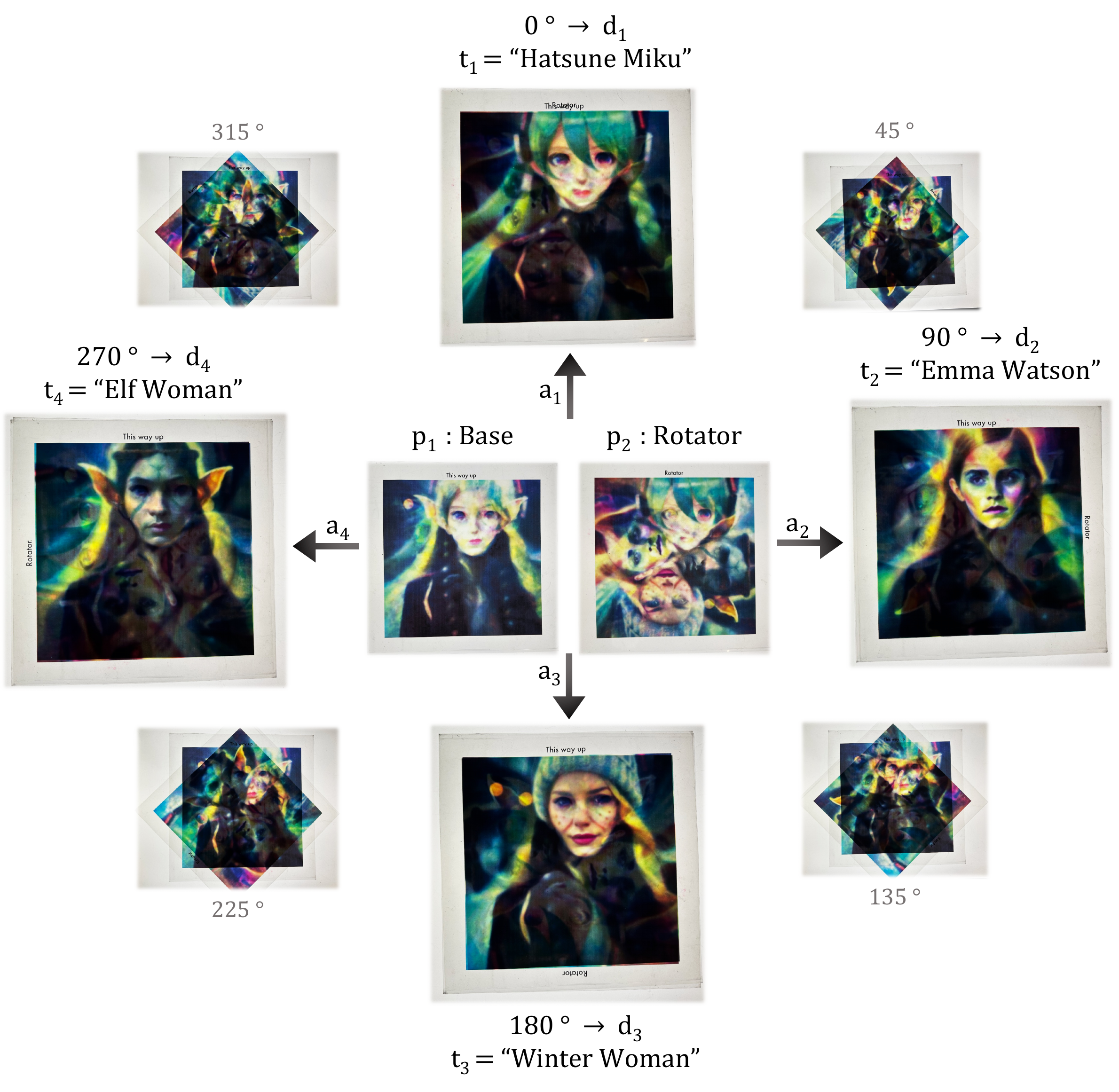}
    \vspace{-2em}
    \caption{This figure shows the rotation overlay illusion arrangement process. \textit{Please note that these are all \underline{real photographs.}} The ``rotator'' image is placed on a ``base'' image over a backlight, both printed out onto transparent sheets. Then, as the rotator spins, we derive four different images. }
    \label{fig:rot_illusion}
    \vspace{-0.5em}
\end{figure}

\item \textbf{Hidden Overlay Illusion} is introduced to push the boundaries of the prime-to-derived relationship, in which four light-filtering prime images, each of which is interpretable on its own, may be merged to obtain a fifth hidden image. Here the modeled view process for the first four derived images is simply the identity function; the view process for the fifth is the product of the four prime images (see \cref{fig:hidden_overlay_animals_photo}).

\end{itemize}
We select these illusion styles to cover varying set cardinalities for prime images and arrangement processes. The arrangement process relevant to each illusion is presented in \cref{tbl:arrangements}. We also present photographs of real-world fabrications for each illusion type in \cref{fig:teaser}, \cref{fig:rot_illusion} and \cref{fig:hidden_overlay_animals_photo}. 

\begin{table}[t]
\centering
\small
\def\arraystretch{1.8}  
\setlength\tabcolsep{0.7em}  
\scalebox{0.86}{
\begin{tabular}{cccc}
    \toprule \rowcolor{Gray}
    Illusion & $n$ & $m$ & $\mathbf{a}$ \\ \midrule
    Flip & 1 & 2 & \makecell{$a_1(\mathbf{p}) = p_1$ \\ $a_2(\mathbf{p}) = \mathrm{rot}(p_2, 180)$} \\ \midrule
    Rotation Overlay & 2 & 4 & $a_j(\mathbf{p}) = p_1 * \mathrm{rot}(p_2, 90j)$ \\ \midrule
    Hidden Overlay & 4 & 5 & \makecell{$a_j(\mathbf{p}) = p_j, j \le 4$ \\ $a_5(\mathbf{p}) = p_1 * p_2 * p_3 * p_4$} \\ \bottomrule
\end{tabular}
}
\vspace{-0.5em}
\caption{This table describes our mathematical models of the Flip, Rotation Overlay, and Hidden Overlay illusions, describing the number of prime images $n$, the number of derived images $m$, and the arrangement operator $\mathbf{a}$ mapping from prime image space $\mathcal{P}^n$ to derived image space $\mathcal{D}^m$. The arrangements in the Flip illusion are simply the identity and a 180 degree rotation. The arrangement operations in the Overlay illusions use a multiplication blend operation to model shining light through multiple transparencies; the result is multiplied by a constant and normalized using $\mathrm{tanh}$ to avoid losing dynamic range.}
\label{tbl:arrangements}
\vspace{-0.5em}
\end{table}


\subsection{Diffusion Illusion Optimization}
\label{sec:losses}
Having selected three diverse illusion styles, we next discuss the process for learning optimal prime images.
Given fully-differentiable operations (also realizable in the physical world) that arrange a set of prime images to produce a derived image, we leverage two types of losses in successive phases to provide suitable alignment signals to the derived images, which in turn would update the prime images. 
In the first phase, we use \textit{Score Distillation Loss}~\cite{Poole2022DreamFusionTU}, a high-fidelity but expensive algorithm that applies a conditional denoising model to the input at every image update step. 
In the second phase, we introduce the complementary \textit{Dream Target Loss}, a faster technique that pulls the derived images towards periodically updated target images.

Given a frozen text-to-image latent diffusion model $\mathcal{F}$~\cite{rombach2022high} which contains a text encoder $\mathcal{F}_t$, an image encoder $\mathcal{F}_e$ and the denoising network $\mathcal{F}_u$, 
we initialize a series of prime images $p_i$ each represented by a Fourier Feature Network with random parameters $\theta_i$. 
Derived images $d_i$ then can be presented by the arrangement process as introduced in \cref{sec:illusions}.
For each derived image $d_i$, a target $t_i$ that describes in natural language the expected visual appearance of its final form is given by the user. 

\subsubsection{Score Distillation Loss}
Score Distillation Loss is a widely-used technique to align images with external conditioning such as textual prompts. 
In essence, Score Distillation Loss ($\mathcal{L}^{\text{SD}}$) randomly selects a timestep $\tau$ of the denoising process, adds noise $\eta_\tau$ proportionate to the timestep $\tau$ to a derived image $d_i$ and applies the denoising process, which is conditioned on corresponding $t_i$, to $d_i + \eta_\tau$ to obtain an estimated noise $\hat{\eta_\tau}$. 
The difference, which we implement as a mean absolute error, between the estimated noise $\hat{\eta_\tau}$ and actual noise $\eta_\tau$ provides a signal for the discrepancy between the derived image $d_i$ and the target description $t_i$ for the derived image. 
This difference is normalized by $\tau$ and then provided as a gradient to the derived image and backpropagated through the arrangement process to the prime image. 
Importantly, this process does not require any backpropagation through the diffusion model. 

In summary, as shown in ~\cref{eq:sd_loss}, score distillation loss provides gradients to optimize the image parameterized by $\theta$, such that iterative updates to the image converge its appearance towards the paired text $t_i$. 

\begin{align}
    \hat{\eta_\tau} &= \mathcal{F}_u(d_i + \eta_\tau, \tau, \mathcal{F}_t(t_i)) \\
    \mathcal{L}_i^{\text{SD}}(t_i, d_i)&=\| \eta_\tau - \hat{\eta_\tau} \|_1
    \label{eq:sd_loss}
\end{align}


\subsubsection{Dream Target Loss}
Dream Target Loss is a novel optimized version of the Score Distillation Loss for circumstances where it is not trivial for prime image(s) to follow the gradients from the Score Distillation Loss. 

\begin{figure}
  \centering
  \includegraphics[width=\linewidth, trim={1cm 1cm 6cm 1cm}, clip]{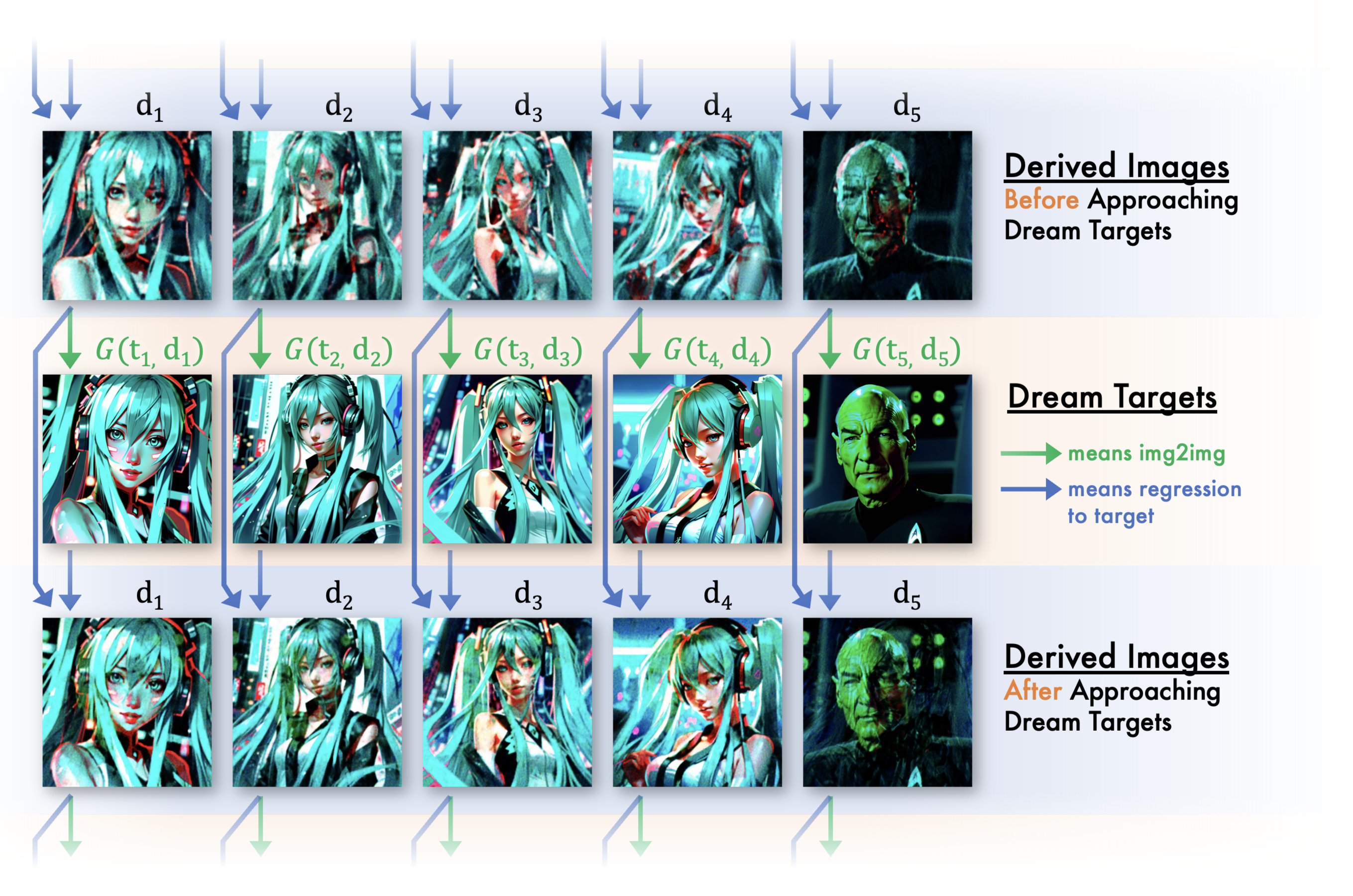}
  \vspace{-2.0em}
  \caption{We depict the dream-target loss above. It is an iterative process, refining derived images using SDEdit to create target images, which the derived images are then regressed to with gradient descent. Note how the derived images look more like the targets after approaching them than before, such as the man's green face.}
  \label{fig:fig-dream-approach}
  \vspace{-0.5em}
\end{figure}

Instead, Dream Target Loss ($\mathcal{L}^{\text{DT}}$) periodically applies a conditional image-to-image process $z_i = \mathcal{G}(t_i, d_i)$ to obtain a target image $z_i$ for each derived image $d_i$, conditioned on the textual prompt $t_i$.
Then we gradually pull each derived image $d_i$ towards its target image $z_i$ using a combination of the structural image similarity loss ($\mathcal{L}_{SSIM}$) and a pixel-wise mean squared error loss ($\mathcal{L}_{2}$). 

Therein, we obtain a joint loss to similarly learn optimal prime images $p_i$ resulting in derived images aligned to each of our target concepts. 
\begin{align}
    z_i &= \mathcal{G}(t_i, d_i)\\
    \mathcal{L}_i^{\text{DT}}(z_i, d_i) &= \mathcal{L}_{\text{SSIM}}(z_i, d_i) + \mathcal{L}_{2}(z_i, d_i)
    \label{eq:img_sim}
\end{align}


An additional feature of the Dream Target Loss relative to the SD variant is that it tends to introduce less noise. 

The total dream target loss is a weighted average across all per derived image loss terms.
\begin{align}
    \mathcal{L}^{\text{DT}} = \sum w_i \mathcal{L}_i^{\text{DT}}
\end{align} where the loss terms are weighted by importance values $w_{1\dots m}$. By default, all $w_i=1$ except in the hidden overlay illusion where the hidden image is prioritized via $w_5=3$.

In practice, for each target image, we optimize the prime image for multiple (i.e. $1000$) steps using the dream target loss. 
Then we repeat the process with the latest prime image so that the target image is updated towards the current derived image for faster convergence (Illustrated in \cref{fig:fig-dream-approach}). 
We implement $\mathcal{G}$ using SDEdit~\cite{meng2022sdedit} where random noise is first added to the input image, and the noisy image is then iteratively denoised conditioned on the text prompt using a frozen diffusion model to generate an output image (see \cref{sec:pseudocode}). 

Note that in both Score Distillation Loss and Dream Target Loss, we propagate gradients to the prime images, updating their parametric representation (i.e. the weights of the MLP Fourier Feature Networks $\theta$), and the diffusion model is kept frozen.

\subsubsection{Visual Prompt}
Optionally, one or more $t_i$ can be given as a specific target image instead of a text prompt --- letting users hide targets such as QR codes or blocks of text. 
In that case, for both phases, the discrepancy between the derived image and the target image is measured using \cref{eq:img_sim}, providing gradients for the prime images.

\subsection{Fabrication}
The flip illusions are trivial to manufacture in real life and need only a printer. The hidden overlay and rotation overlay illusions are created by printing their prime images on overhead display sheets on a color laser printer, before being laminated to protect them from scratches.  
With a strong enough backlight, the hidden overlays and rotation overlay illusions can be performed on regular pieces of paper as well.



\section{Experiments}
In this section, we evaluate our framework presenting qualitative visualizations and quantitative metrics. 

\subsection{Qualitative Evaluation}
\begin{figure}
    \centering
    \includegraphics[width=.7\linewidth, trim={0 0 0 1cm}, clip]{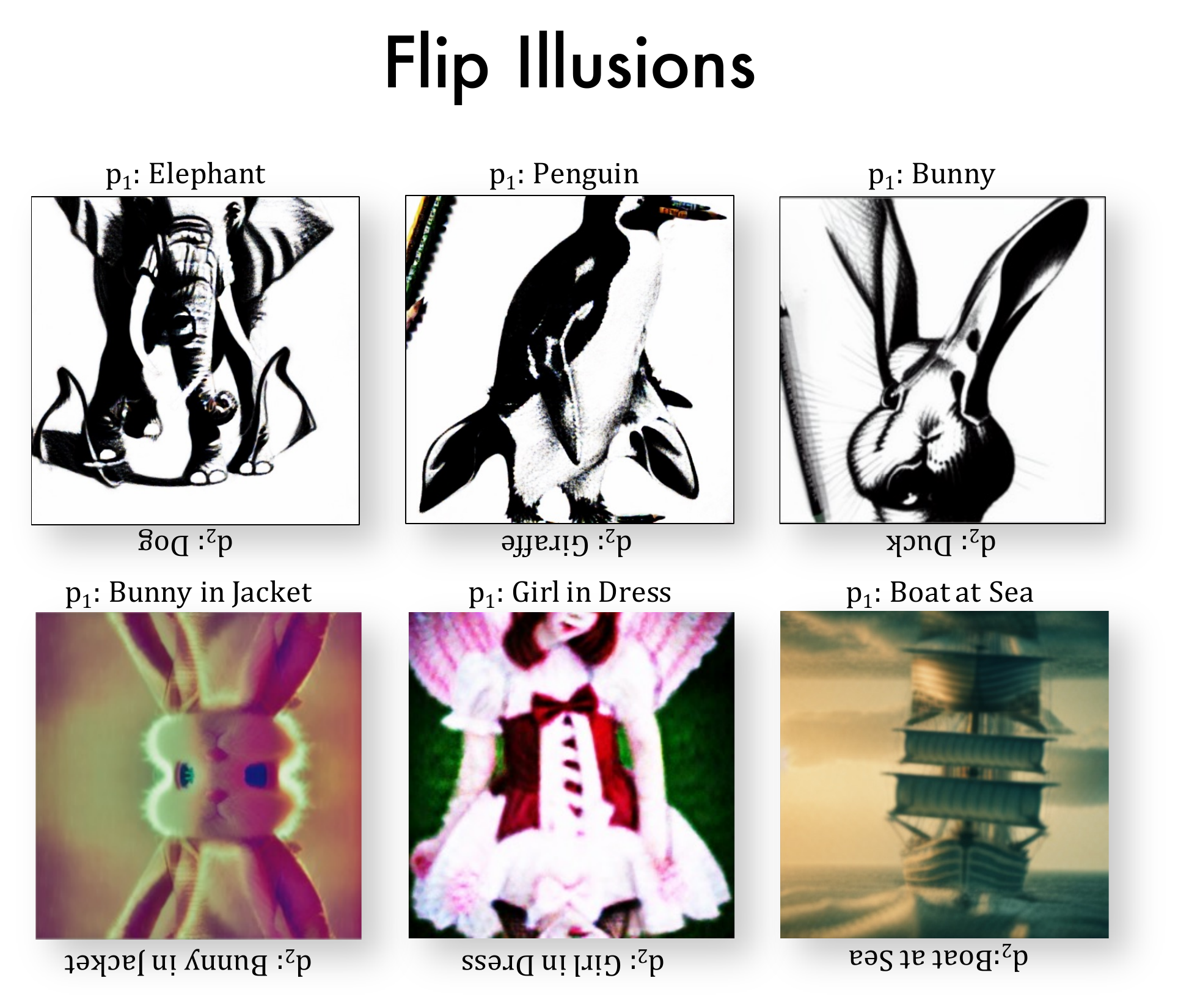}
    \vspace{-0.5em}
    \caption{\textbf{Flip Illusion Examples:} Please view these images upside-down as well as right-side-up to see two different subjects.  Note: In this illusion,  $d_1 = p_1$}
    \label{fig:galflip}
    \vspace{-0.5em}
\end{figure}

\begin{figure}
    \centering
    \includegraphics[width=0.98\linewidth,trim={0cm 0cm 1cm 0cm}, clip]{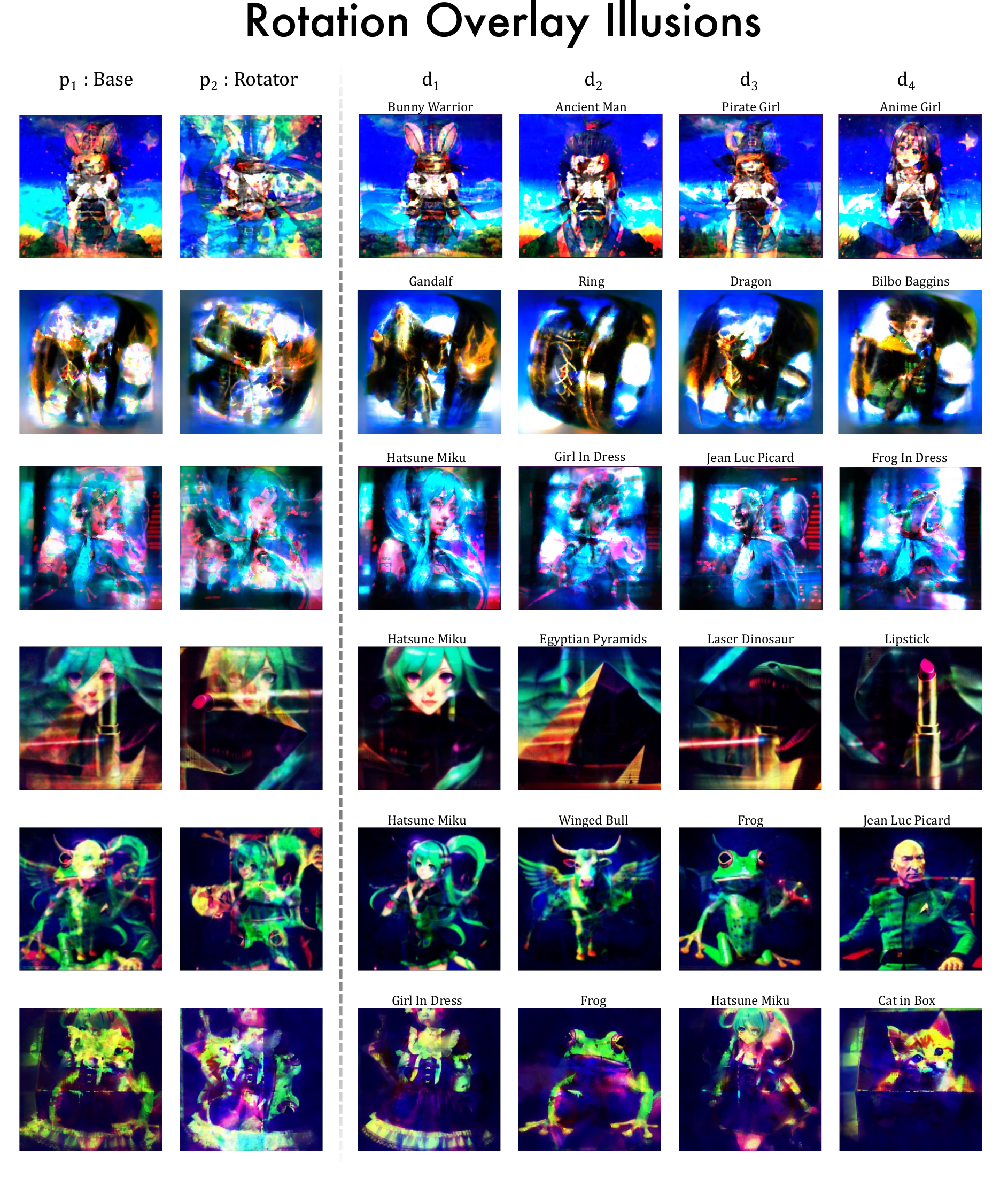}
    \vspace{-0.5em}
    \caption{\textbf{Rotation Overlay Examples:}  On the left are the two prime images $p_1$, $p_2$, and on the right are the four derived images $d_{1\dots 4}$ that are obtained by taking the product of the primes, simulation of them overlaid on a backlight.}
    \vspace{-0.5em}
    \label{fig:galrot}
\end{figure}

\begin{figure}
    \centering
    \includegraphics[width=0.96\linewidth, trim={0 0.95cm 0 0}, clip]{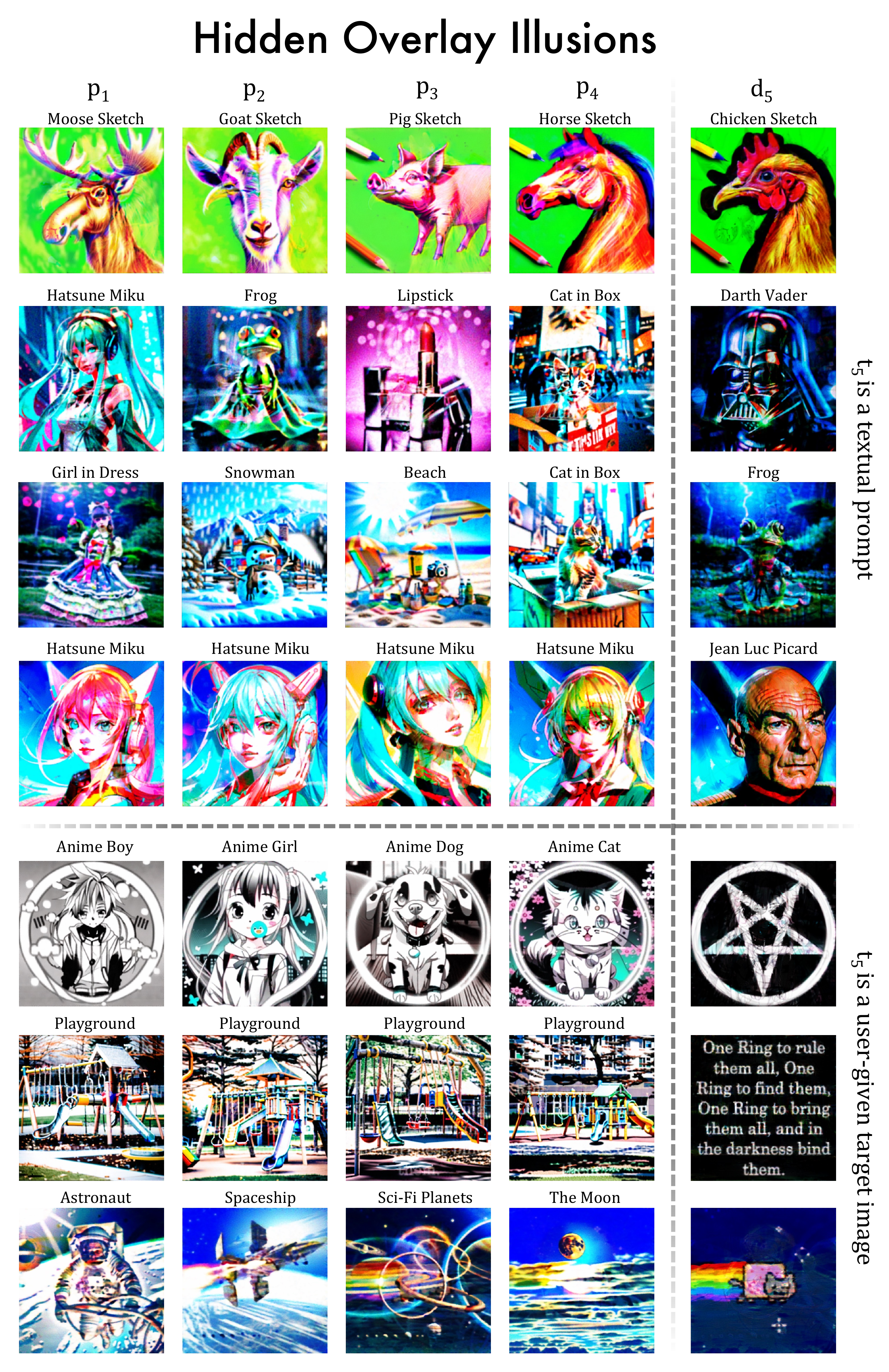}
    \vspace{-0.5em}
    \caption{\textbf{Hidden Overlay Examples:} On the left are the four prime images $p_1$, $p_2$, $p_3$, $p_4$ and on the right is the derived image $d_5 = p_1 \cdot p_2 \cdot p_3 \cdot p_4$, which simulates overlaying them over a backlight. Note: In this illusion,  $d_i = p_i$ for $i\in 1\dots 4$}
    \vspace{-0.5em}
    \label{fig:galhid}
\end{figure}

We illustrate randomly selected example outputs of our Diffusion Illusions framework. Visualizations for our three selected illusion styles, Flip Illusion, Rotation Overlay Illusion, and Hidden Overlay Illusion are presented in \cref{fig:galflip},
\cref{fig:galrot}, and \cref{fig:galhid} respectively. 
For more interactive examples, please refer to the project website \projurl

\subsection{Quantitative Evaluation}
\label{sec:numbers}
Next, we quantitatively benchmark the Hidden Overlay Illusion generated by the variants of Diffusion Illusion in multiple aspects and demonstrate the generalization ability and robustness of the proposed framework. Please check  Appendix~\ref{sec:more_eval} as well for other illusions and more details.

\noindent\textbf{Image Generation Protocol} \quad We design a pipeline that constructs diverse textual prompts randomly and automatically. 
The pipeline relies on two sets of textual prompts. 
The first set $T^s$ is of sentences where each sentence describes a unique art style of an image and contains one \textit{subject} token representing the potential subject of the sentence. 
The second set $T^o$ is of different subjects like `dog', `cat', `car', and so on.
When generating images with a specific style $t^s \in T^s$, we uniformly sample five unique subjects $t^o_i$ where $i \in\{1, \ldots, 5\}$ from $T^o$.
Then we substitute the \textit{subject} token in $t^s$ with $t^o_i$ to construct the textual prompt $t_i$.
Finally, $t_1, \ldots, t_5$ is used to guide the generation of derived images.

For a full evaluation, the whole pipeline is repeated for $N$ times per style $t^s$ to generate $N$ groups of illusion images.
In practice, we set $|T^s| = 4$, $T_o$ is the set of all object classes except `person' in PASCAL VOC~\cite{everingham2010pascal} ($|T^o|=19$), and $N=64$. 
Please refer to the Appendix~\ref{sec:more_eval} for the complete list of subjects and styles.

\noindent\textbf{Evaluation Metrics} \quad Inspired by recent works on diffusion model evaluation~\cite{yeh2023navigating, lee2023holistic}, we measure the following properties of the derived images:

\begin{itemize}
    \item \textit{Controllability:} how well the generated images align with the textual prompts. For each generated image and its corresponding textual prompt, we measure the \textit{average cosine similarity} between the image embedding and the text embedding, extracted from a pretrained CLIP~\cite{radford2021learning} model.
    \item \textit{Diversity:} the variety of generated images conditioned on the same prompt. For images generated by the same textual prompt, we calculate two \textit{Venti scores}~\cite{friedman2022vendi} independently based on two visual embeddings: the \texttt{[CLS]} embeddings of DINOv2~\cite{oquab2023dinov2} and CLIP visual embeddings (see Appendix). 
    \item \textit{Aesthetics:} the assessment of an image's visual appeal and artistic quality. For each image, we utilize AVA LAION-Aesthetics Predictor V2, which is pretrained on AVA~\cite{murray2012ava} dataset, to estimate an aesthetics score range from 0 to 10. 
    
\end{itemize}

In addition, we study a new property ~\textit{Independence} specifically for the illusion scenario. 
Intuitively, each image is expected to stick to its corresponding textual prompt while not being distracted by other textual prompts in the same group.
Such property is named as \textit{Independence}, which is different from \textit{Controllability} because independence is designed to reflect not only the similarity between an image and its corresponding textual prompt but also the \textit{dissimilarity} between the image and the textual prompts for other images. 
In other words, this property focuses on how well the prime images can `hide' the overlay image or how challenging it will be for people to infer the overlay image from a single prime image and vice versa. 

\begin{itemize}
\item \emph{Independence Score:} Therefore, we propose a new metric Independence Score to reflect such property. Consider a set of $m$ derived images, denoted as $\{d_1, d_2, \ldots, d_m\}$, along with their corresponding textual prompts $\{t_1, t_2, \ldots, t_m\}$. Initially, we extract the visual embeddings $v_i=f_v(d_i)$ and text embeddings $e_j=f_t(t_j)$ using the visual encoder $f_v$ and the text encoder $f_t$ from a pretrained CLIP~\cite{radford2021learning} model respectively. Subsequently, we compute the cosine similarity $k_{ij}=\text{CosineSimilarity}(v_i, e_j)$ between any visual and text embeddings $v_i$ and $e_j$. The results are assembled into a matrix $K$, where $k_{ij}$ is put in the $i$-th row and $j$-th column. The Independence Score $S_\mathrm{IS}$ is calculated by the following equations.
\begin{align}
K_0 &= \mathrm{Softmax}(K / \tau, 0) \\ 
K_1 &= \mathrm{Softmax}(K / \tau, 1) \\ 
S_\mathrm{IS} & := \min(\mathrm{diag}(K_0) \cup \mathrm{diag}(K_1)) 
\end{align}
where $\tau=0.05$ is a temperature constant, $\text{Softmax}(\cdot, l)$ stands for softmax operation along $l$-th dimension and $\text{diag}(\cdot)$ presents a set of the diagonal elements of $(\cdot)$. 
$S_\mathrm{IS}$ is designed to become higher when all images $d_i$ align best with their corresponding textual prompts compared with other textual prompts.
\end{itemize}

\noindent\textbf{Methods} \quad The baseline method of our experiments is a vanilla SDXL generating target images with corresponding textual prompts independently for one step using score distillation loss. We benchmark four variants of our methods named A, B, C, and D. Method C is our default method. It involves 500 steps of score distillation loss followed by 8 steps of dream target loss and applies relative weights [1,1,1,1,3] respectively - which prioritizes the quality-derived hidden image over its constituent primes. In addition, Method A uses Stable Diffusion 1.5 instead of SDXL, which is used by all other methods. Method B uses equal weights for all derived images, using weights [1,1,1,1,1] respectively. Lastly, method D uses 4000 steps of score distillation loss followed by 1 step of dream target loss for smoothness, to evaluate the ability of score distillation loss alone in this task.
For fairness, all methods were constrained to run in a 15-minute time window on a single NVIDIA A100 GPU.

\begin{figure}[h]
  \centering
  \includegraphics[width=1\linewidth]{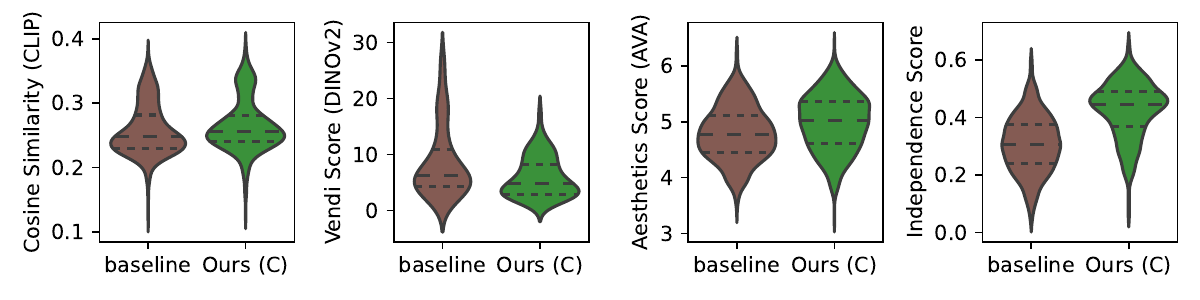}
  \vspace{-0.5em}
  \caption{Comparison of multiple score distributions. Refer to axes for metrics. Our framework clearly outperforms the baseline for all metrics except diversity (Vendi Score). We argue the additional constraints intrinsic to our task (of generating illusions) contributed to reduced diversity.}
  \label{fig:big-baseline-fig}
\end{figure}

\noindent\textbf{Results} \quad For all metrics, we report the score distributions achieved by our default method and the baseline in \cref{fig:big-baseline-fig}. 

Our method significantly outperforms the baseline in all metrics except the Vendi Score, which is expected because, for our method, there are more constraints from the derived images applied during the generation process.

\begin{figure}[h]
  \centering
  \includegraphics[width=\linewidth]{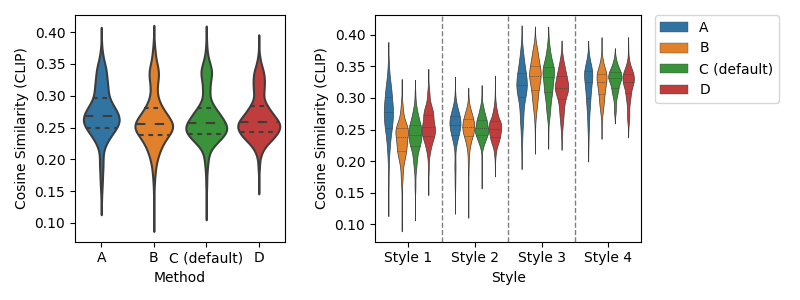}
  \includegraphics[width=\linewidth]{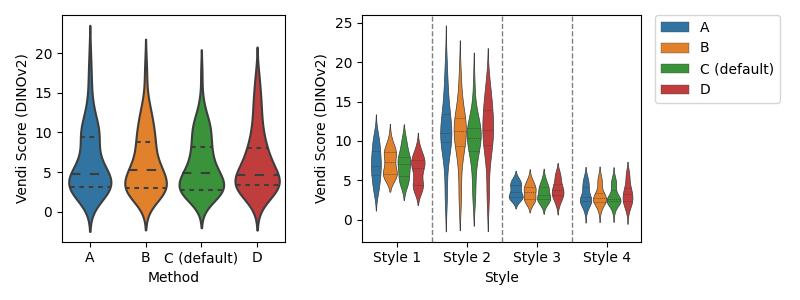}
  \includegraphics[width=\linewidth]{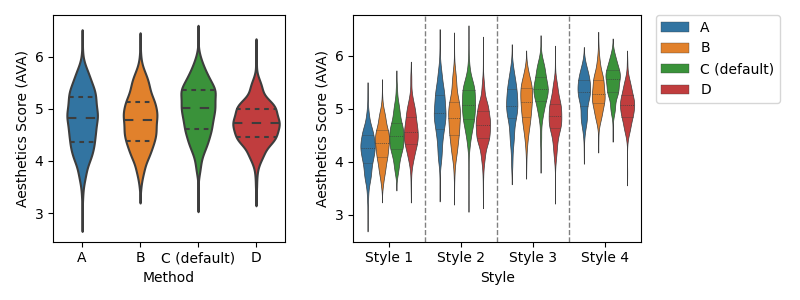}
  \includegraphics[width=\linewidth]{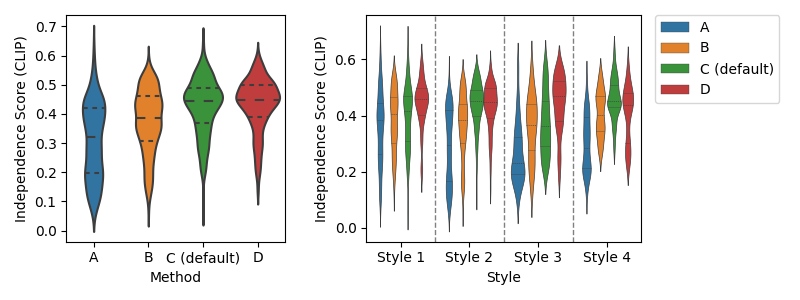}
  \vspace{-0.5em}
  \caption{Score distributions over methods (left) and styles (right). A, B, C, D stands for four variants of our method. Results indicate the significance of prompts for illusion generation.}
  \label{fig:big-score-fig}
\end{figure}

The score distributions of four variants of our method are presented in \cref{fig:big-score-fig}. Each row of \cref{fig:big-score-fig} presents two metrics. The subfigures on the left-hand side show the overall performance of a specific method. 
In general, all methods perform similarly well in terms of Controllability (Cosine Similarity) and Diversity (Vendi Score) (the first two rows in \cref{fig:big-score-fig}). Method C shows significant advantages in Aesthetics (Aesthetics Score) and Methods C and D achieve relatively higher Independence Score. 

A detailed look at different art styles is presented on the right-hand side of each row of \cref{fig:big-score-fig}, where different metrics respond diversely to different art styles.
Controllability (Cosine Similarity) prefers Style 3 and Style 4 while the Diversity (Vendi Score) prefers Style 2. 
The Aesthetics Score and Independence Score are generally robust to the different styles.
However, the Aesthetics Score prefers Style 4 slightly more than Style 1.

In conclusion, the prompts used are far more important than the chosen implementation. There is no clear one-size-fits-all method indicated by our quantitative evaluations, however, we observe that depending on the art styles and subjects used, a different method will be optimal. 
One should carefully pick up a method when generating illusions in a specific art style.
A further study on subjects is available in the Appendix.

\subsection{Discussions}
\label{sec:discussion}
In this section, we discuss several observations that may inspire future investigation.

\noindent\textit{Q1: Can Diffusion Illusion yield better images when running for a longer time?} 

\begin{figure}[h]
  \centering
  \includegraphics[width=0.4\textwidth]{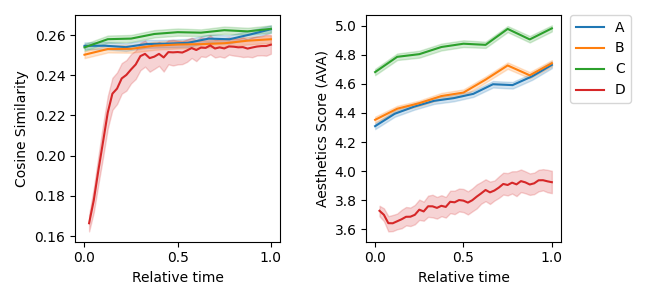}
  \caption{CLIP Cosine Similarity (left) and Aesthetics Score (right) increase when optimizing for a longer time.}
  \label{fig:fig-grad-increase}
\end{figure}

\noindent Yes. \cref{fig:fig-grad-increase} presents the trend of Controllability (Cosine Similarity) and Aesthetics (Aesthetics Score) as the images used in \cref{sec:numbers} are getting optimized. 
The term `relative time' is employed to denote the progression of wall-clock time during the optimization process. 
A relative time value of 0 means the beginning of optimization, whereas a value of 1 marks its conclusion. 
\cref{fig:fig-grad-increase} reveals a notable trend: there is a consistent increase in metrics as the optimization process advances.

\noindent\textit{Q2: Is Independence Score a qualitatively valid metric?}

\begin{figure}[h]
  \centering
  \includegraphics[width=\linewidth]{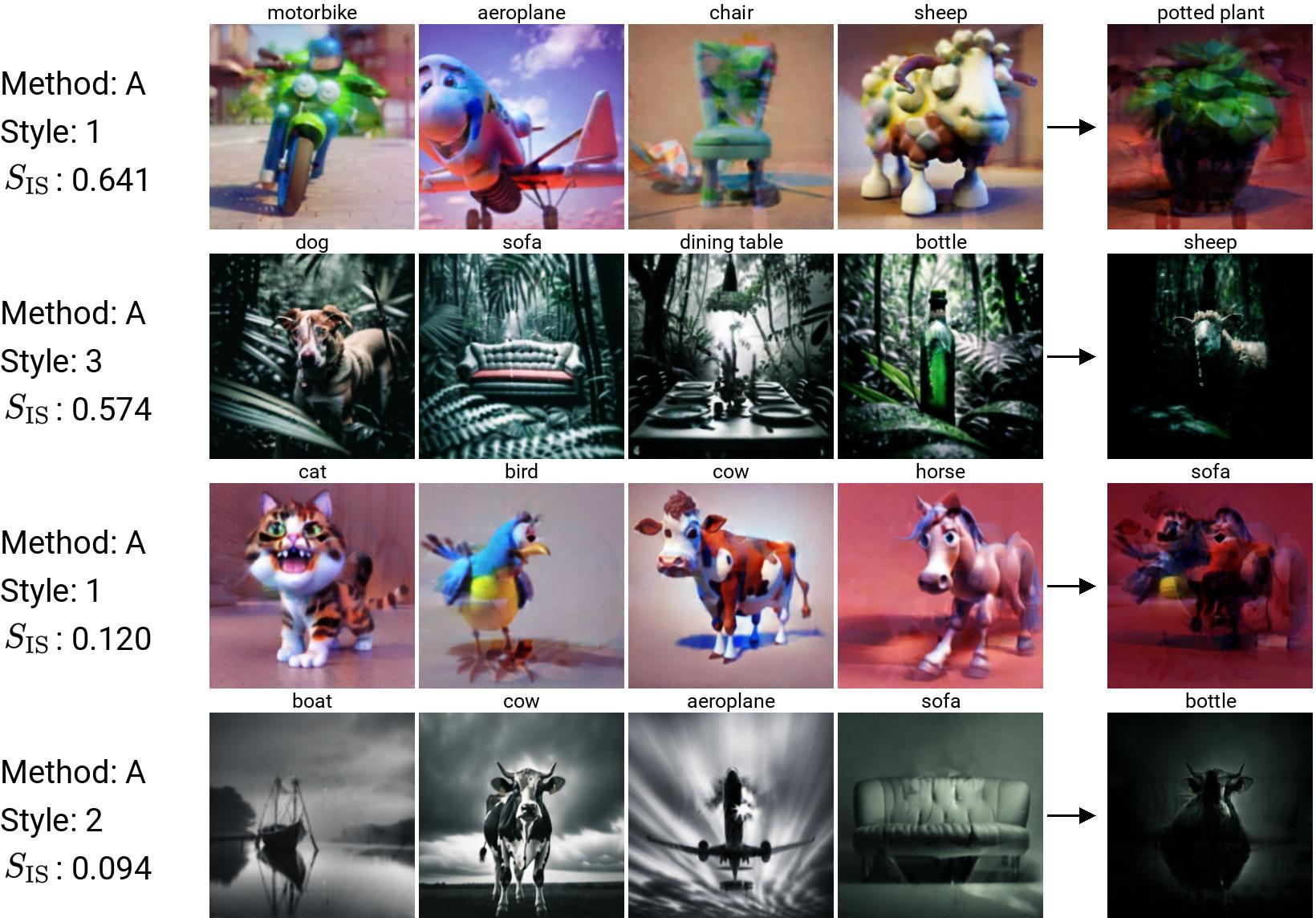}
  \caption{Examples with diverse independence scores}
  \label{fig:fig-sis}
\end{figure}

\noindent Generally yes. \cref{fig:fig-sis} shows four illusions randomly selected with diverse independence scores. For each row, the subject of each image is listed above, and the method, style, and independence scores are listed on the left-hand side. 
The four images grouped in the middle are prime images and they derive the overlay image on the right-hand side.
For the first two examples where the independence score is relatively high, each image aligns with its corresponding textual prompt. 
However, for the third example, the overlay image is not closely related to the subject `sofa', resulting in a lower independence score. 
Furthermore, in the last example of \cref{fig:fig-sis}, the overlay image visually biases more towards `cow' instead of `bottle', leading to the lowest independence score.

\noindent\textit{Q3: What are the reasons to use Fourier Features Network?} 

\begin{figure}[h]
  \centering
  \includegraphics[width=\linewidth]{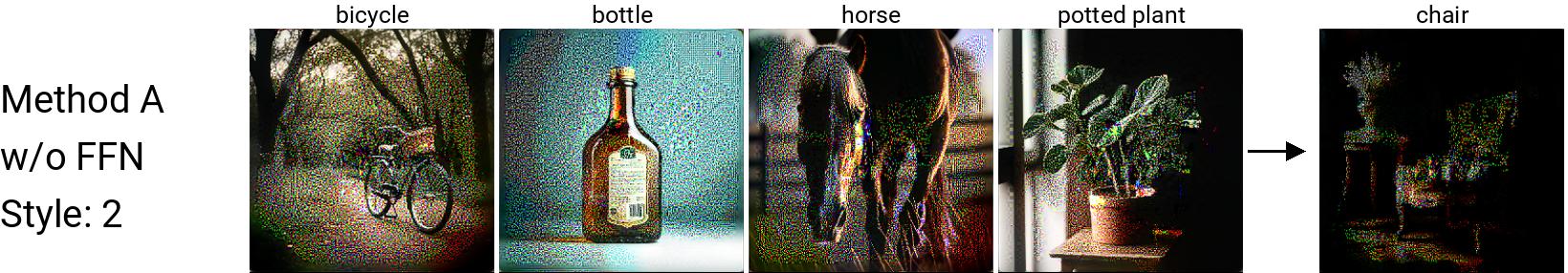}
  \caption{A Hidden Overlay image with prime images optimized directly in pixel space. While high-frequency encoding of the hidden image results in less perceivable interference in each individual image, it results in a brittle illusion that is disrupted without pixel-perfect printing and alignment.}
  \label{fig:fig-noisy}
\end{figure}

\noindent Earlier experiments optimizing prime images directly in pixel space resulted in information being encoded at very high frequencies and requiring pixel-perfect alignment to generate the intended derived images (see \cref{fig:fig-noisy}). While the result was pleasing when viewed digitally, it was impractical for real-world illusions. Motivated by previous arguments \cite{Burgert2022PeekabooTT, Burgert2022}, we elect to use Fourier Features Network \cite{tancik2020fourier} based parametric image representations. 

\section{Conclusion}
In this paper, we establish the formal definition of the problem of generating illusions and introduce Diffusion Illusions, a versatile pipeline designed for the generation of a diverse array of illusions. 
Complemented by comprehensive experiments conducted across multiple facets, we verify the effectiveness of our proposed method qualitatively and quantitatively.
We also successfully fabricate the prime images in the real world.
Other areas to explore include more types of illusion generation and creative ways to take advantage of diffusion models. 

\vspace{2em}
\noindent
\textbf{Limitations}  \quad
The main limitation of our framework is the relatively high inference time required for generating illusions. While our framework improves over plain score distillation in terms of inference time, we are still slow. Improving the speed of illusion generation frameworks such as ours presents an interesting future direction. 
We note that contemporary work has already explored ways to minimize this inference time. 

\begin{figure}[t]
  \centering
  \includegraphics[width=1\linewidth, trim={1cm 1cm 1cm 1cm}, clip]{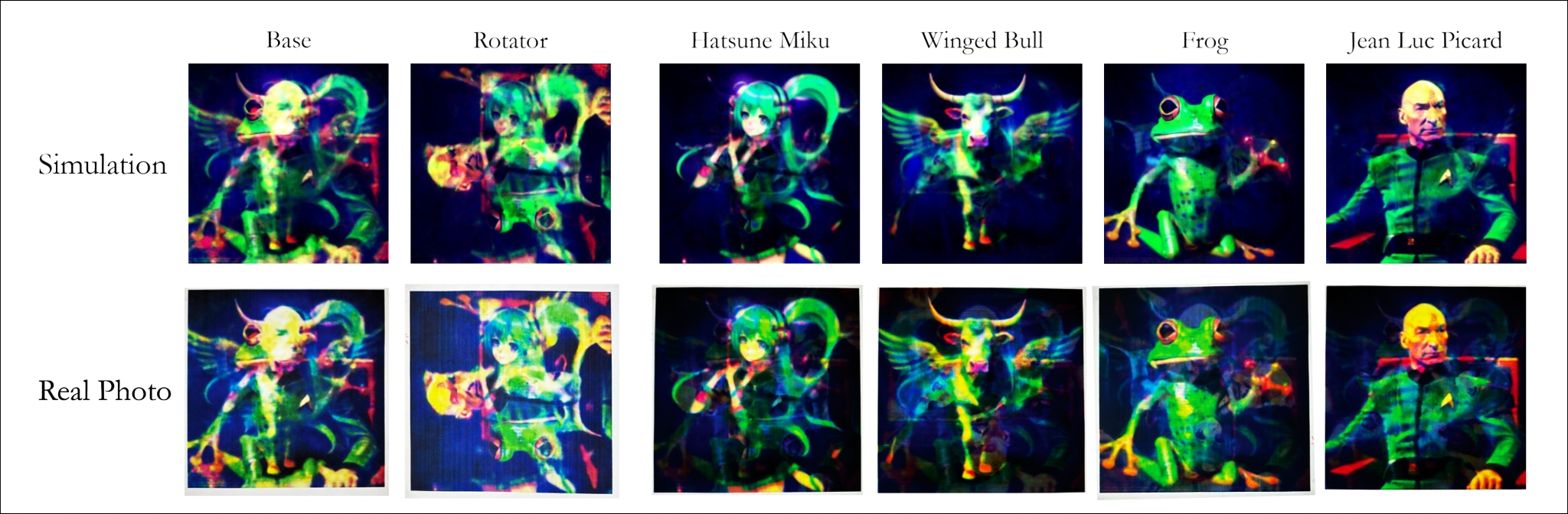}
  \caption{The colors shift after printing out Rotation Overlay Illusion images. First row: digital copy of the images and the overlay simulation. Second row: real-world photos of the printed images.}
  \label{fig:simreal2}
\end{figure}

\begin{figure}[t]
  \centering
  \includegraphics[width=1\linewidth]{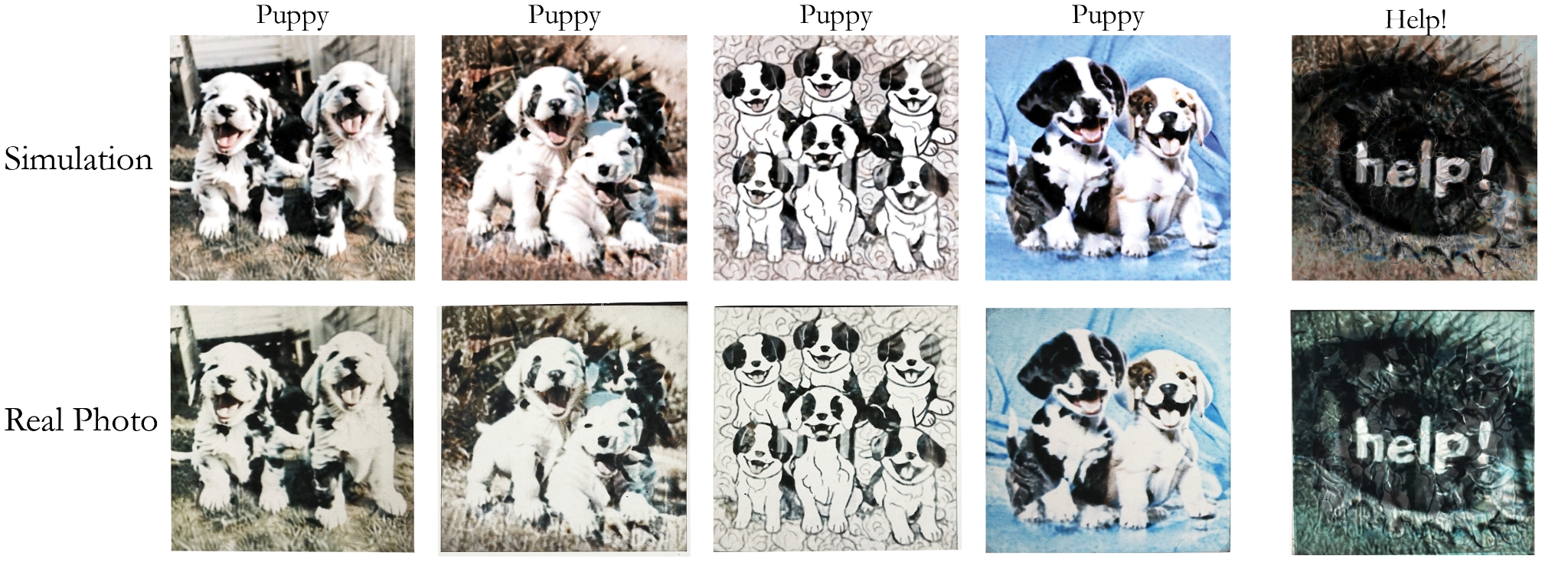}
  \caption{The colors shift after printing out Hidden Overlay Illusion images. First row: digital copy of the images and the overlay simulation. Second row: real-world photos of the printed images.}
  \label{fig:simreal}
\end{figure}

Furthermore, 
the effectiveness of visual illusion in the real world may vary a lot due to the errors introduced in the printing process. \cref{fig:simreal2} and \cref{fig:simreal} present the effect of the color shifts when printing the images.

Other limitations include biases contained in our models (discussed in detail under the ethics statement). 

\vspace{1em}
\noindent
\textbf{Reproducibility Statement}  \quad
Our work builds off open-source models whose pre-trained weights are publicly available. Our framework simply performs inference time optimizations to generate illusions. In our paper, we detail all specifics of our implementation (including PyTorch style pseudo-code) necessary to generate such illusions. Our code (and all material necessary to replicate results in paper) will be released publicly. 

\vspace{1em}
\noindent
\textbf{Ethics Statement}  \quad
A main ethical concern for any generative art model is that it will reduce the demand for human artists in its domain. Generating optical illusion artwork is a very difficult artistic task, and there are few artists that attempt it. Thus, the genre of illusions is currently relatively small and there is limited demand for illusions at present. Diffusion Illusions makes the generation of optical illusions accessible to the general public, making illusions more accessible to the layperson. We believe that, if anything, Diffusion Illusions and related works are likely to increase interest in illusions and the demand for human-created illusions as a result.
Secondly, our experiments utilize Stable Diffusion 1.5 and Stable Diffusion-XL models, and thus our reference implementation of the Diffusion Illusions pipeline will replicate any biases contained within these models. These models are trained on the LAION-2B(en) and LAION-5B datasets, and may over-represent English-language or Western content. The Stable Diffusion 1.5 and Stable Diffusion-XL models are intended for research purposes only, and thus our reference implementation should also be used exclusively for research and informative purposes. Some recent models, including DeepFloyd, are licensed for limited production use and our pipeline easily generalizes to them; however, they have higher system requirements.

\vspace{1em}
\noindent
\textbf{Contributions}  \quad
RB led the project, conceived the prime image / derived image illusion relationship, invented the classes of hidden and rotation overlay illusions, and designed \& implemented the Diffusion Illusions pipeline.
XL designed and performed all quantitative evaluation experiments.
AL formalized and wrote the Illusion problem statement and contributed to paper writing.
KR discussed multiple aspects of the project, supported designing a prior framework (Peekaboo~\cite{Burgert2022PeekabooTT}) important for building our setup, and contributed to paper writing.
MR supervised the project, advised on research direction, and discussed all aspects of the project.  

\vspace{1em}
\noindent \textbf{Acknowledgements} \quad
We would like to thank Brian Price and Jinghuan Shang for helpful discussions about this paper, and Jongwoo Park for both his discussions as well as posing for the flip illusion photo in \cref{fig:teaser}. This material is based upon work supported by the National Science Foundation Graduate Research Fellowship under Grant No. 2234683.


{
    \small
    \bibliographystyle{ieeenat_fullname}
    \bibliography{main}

\begin{thebibliography}{44}
\providecommand{\natexlab}[1]{#1}
\providecommand{\url}[1]{\texttt{#1}}
\expandafter\ifx\csname urlstyle\endcsname\relax
  \providecommand{\doi}[1]{doi: #1}\else
  \providecommand{\doi}{doi: \begingroup \urlstyle{rm}\Url}\fi

\bibitem[Benny et~al.(2020)Benny, Galanti, Benaim, and Wolf]{Benny2020EvaluationMF}
Yaniv Benny, Tomer Galanti, Sagie Benaim, and Lior Wolf.
\newblock Evaluation metrics for conditional image generation.
\newblock \emph{International Journal of Computer Vision}, 129:\penalty0 1712 -- 1731, 2020.

\bibitem[Betzalel et~al.(2022)Betzalel, Penso, Navon, and Fetaya]{Betzalel2022ASO}
Eyal Betzalel, Coby Penso, Aviv Navon, and Ethan Fetaya.
\newblock A study on the evaluation of generative models.
\newblock \emph{ArXiv}, abs/2206.10935, 2022.

\bibitem[Boring(1930)]{boring_new_1930}
E.~G. Boring.
\newblock A new ambiguous figure.
\newblock \emph{The American Journal of Psychology}, 42:\penalty0 444--445, 1930.
\newblock Place: US Publisher: Univ of Illinois Press.

\bibitem[Burgert et~al.(2022{\natexlab{a}})Burgert, Ranasinghe, Li, and Ryoo]{Burgert2022PeekabooTT}
Ryan Burgert, Kanchana Ranasinghe, Xiang Li, and Michael~S. Ryoo.
\newblock Peekaboo: Text to image diffusion models are zero-shot segmentors.
\newblock \emph{ArXiv}, abs/2211.13224, 2022{\natexlab{a}}.

\bibitem[Burgert et~al.(2022{\natexlab{b}})Burgert, Shang, Li, and Ryoo]{Burgert2022}
Ryan Burgert, Jinghuan Shang, Xiang Li, and Michael Ryoo.
\newblock Neural neural textures make sim2real consistent.
\newblock In \emph{Proceedings of the 6th Conference on Robot Learning}, 2022{\natexlab{b}}.

\bibitem[Darcet et~al.(2023)Darcet, Oquab, Mairal, and Bojanowski]{darcet2023vision}
Timoth{\'e}e Darcet, Maxime Oquab, Julien Mairal, and Piotr Bojanowski.
\newblock Vision transformers need registers.
\newblock \emph{arXiv preprint arXiv:2309.16588}, 2023.

\bibitem[Dhariwal and Nichol(2021)]{diffusionbeatsgans}
Prafulla Dhariwal and Alex Nichol.
\newblock Diffusion models beat gans on image synthesis.
\newblock \emph{ArXiv}, abs/2105.05233, 2021.

\bibitem[Dosovitskiy et~al.(2020)Dosovitskiy, Beyer, Kolesnikov, Weissenborn, Zhai, Unterthiner, Dehghani, Minderer, Heigold, Gelly, et~al.]{dosovitskiy2020image}
Alexey Dosovitskiy, Lucas Beyer, Alexander Kolesnikov, Dirk Weissenborn, Xiaohua Zhai, Thomas Unterthiner, Mostafa Dehghani, Matthias Minderer, Georg Heigold, Sylvain Gelly, et~al.
\newblock An image is worth 16x16 words: Transformers for image recognition at scale.
\newblock \emph{arXiv preprint arXiv:2010.11929}, 2020.

\bibitem[Everingham et~al.(2010)Everingham, Van~Gool, Williams, Winn, and Zisserman]{everingham2010pascal}
Mark Everingham, Luc Van~Gool, Christopher~KI Williams, John Winn, and Andrew Zisserman.
\newblock The pascal visual object classes (voc) challenge.
\newblock \emph{International journal of computer vision}, 88:\penalty0 303--338, 2010.

\bibitem[Friedman and Dieng(2022)]{friedman2022vendi}
Dan Friedman and Adji~Bousso Dieng.
\newblock The vendi score: A diversity evaluation metric for machine learning.
\newblock \emph{arXiv preprint arXiv:2210.02410}, 2022.

\bibitem[Geng et~al.(2023)Geng, Park, and Owens]{geng2023visualanagrams}
Daniel Geng, Inbum Park, and Andrew Owens.
\newblock Visual anagrams: Generating multi-view optical illusions with diffusion models.
\newblock \emph{arXiv:2311.17919}, 2023.

\bibitem[Hofstadter(1985)]{hofstadter1985meta}
Douglas~R Hofstadter.
\newblock Metafont, metamathematics, and metaphysics: Comments on donald knuth's article ``the concept of a meta-font''.
\newblock \emph{Metamagical themas: Questing for the essence of mind and pattern}, pages 274--278, 1985.

\bibitem[Hsiao et~al.(2018)Hsiao, Huang, and Chu]{hsiao_multi-view_2018}
Kai-Wen Hsiao, Jia-Bin Huang, and Hung-Kuo Chu.
\newblock Multi-view wire art.
\newblock \emph{ACM Transactions on Graphics}, 37\penalty0 (6):\penalty0 1--11, 2018.

\bibitem[Jastrow(1899)]{jastrow1899mind}
Joseph Jastrow.
\newblock The mind's eye.
\newblock \emph{Popular Science Monthly}, pages 299--312, 1899.

\bibitem[Lee et~al.(2023{\natexlab{a}})Lee, Yasunaga, Meng, Mai, Park, Gupta, Zhang, Narayanan, Teufel, Bellagente, Kang, Park, Leskovec, Zhu, Fei-Fei, Wu, Ermon, and Liang]{lee_holistic_2023}
Tony Lee, Michihiro Yasunaga, Chenlin Meng, Yifan Mai, Joon~Sung Park, Agrim Gupta, Yunzhi Zhang, Deepak Narayanan, Hannah~Benita Teufel, Marco Bellagente, Minguk Kang, Taesung Park, Jure Leskovec, Jun-Yan Zhu, Li Fei-Fei, Jiajun Wu, Stefano Ermon, and Percy Liang.
\newblock Holistic {Evaluation} of {Text}-{To}-{Image} {Models}, 2023{\natexlab{a}}.
\newblock arXiv:2311.04287 [cs].

\bibitem[Lee et~al.(2023{\natexlab{b}})Lee, Yasunaga, Meng, Mai, Park, Gupta, Zhang, Narayanan, Teufel, Bellagente, et~al.]{lee2023holistic}
Tony Lee, Michihiro Yasunaga, Chenlin Meng, Yifan Mai, Joon~Sung Park, Agrim Gupta, Yunzhi Zhang, Deepak Narayanan, Hannah~Benita Teufel, Marco Bellagente, et~al.
\newblock Holistic evaluation of text-to-image models.
\newblock \emph{arXiv preprint arXiv:2311.04287}, 2023{\natexlab{b}}.

\bibitem[Liu et~al.(2023{\natexlab{a}})Liu, Li, Li, and Lee]{liu2023improved}
Haotian Liu, Chunyuan Li, Yuheng Li, and Yong~Jae Lee.
\newblock Improved baselines with visual instruction tuning.
\newblock \emph{arXiv preprint arXiv:2310.03744}, 2023{\natexlab{a}}.

\bibitem[Liu et~al.(2023{\natexlab{b}})Liu, Li, Wu, and Lee]{liu2023visual}
Haotian Liu, Chunyuan Li, Qingyang Wu, and Yong~Jae Lee.
\newblock Visual instruction tuning.
\newblock \emph{arXiv preprint arXiv:2304.08485}, 2023{\natexlab{b}}.

\bibitem[Meng et~al.(2022)Meng, He, Song, Song, Wu, Zhu, and Ermon]{meng2022sdedit}
Chenlin Meng, Yutong He, Yang Song, Jiaming Song, Jiajun Wu, Jun-Yan Zhu, and Stefano Ermon.
\newblock Sdedit: Guided image synthesis and editing with stochastic differential equations, 2022.

\bibitem[Murray et~al.(2012)Murray, Marchesotti, and Perronnin]{murray2012ava}
Naila Murray, Luca Marchesotti, and Florent Perronnin.
\newblock Ava: A large-scale database for aesthetic visual analysis.
\newblock In \emph{2012 IEEE conference on computer vision and pattern recognition}, pages 2408--2415. IEEE, 2012.

\bibitem[Nichol et~al.(2022)Nichol, Dhariwal, Ramesh, Shyam, Mishkin, McGrew, Sutskever, and Chen]{Nichol2022GLIDETP}
Alex Nichol, Prafulla Dhariwal, Aditya Ramesh, Pranav Shyam, Pamela Mishkin, Bob McGrew, Ilya Sutskever, and Mark Chen.
\newblock Glide: Towards photorealistic image generation and editing with text-guided diffusion models.
\newblock In \emph{ICML}, 2022.

\bibitem[Nicholls et~al.(2018)Nicholls, Churches, and Loetscher]{nicholls_perception_2018}
Michael E.~R. Nicholls, Owen Churches, and Tobias Loetscher.
\newblock Perception of an ambiguous figure is affected by own-age social biases.
\newblock \emph{Scientific Reports}, 8:\penalty0 12661, 2018.

\bibitem[Oliva et~al.(2006)Oliva, Torralba, and Schyns]{oliva2006hybrid}
Aude Oliva, Antonio Torralba, and Philippe~G Schyns.
\newblock Hybrid images.
\newblock \emph{ACM Transactions on Graphics (TOG)}, 25\penalty0 (3):\penalty0 527--532, 2006.

\bibitem[Oquab et~al.(2023)Oquab, Darcet, Moutakanni, Vo, Szafraniec, Khalidov, Fernandez, Haziza, Massa, El-Nouby, et~al.]{oquab2023dinov2}
Maxime Oquab, Timoth{\'e}e Darcet, Th{\'e}o Moutakanni, Huy Vo, Marc Szafraniec, Vasil Khalidov, Pierre Fernandez, Daniel Haziza, Francisco Massa, Alaaeldin El-Nouby, et~al.
\newblock Dinov2: Learning robust visual features without supervision.
\newblock \emph{arXiv preprint arXiv:2304.07193}, 2023.

\bibitem[Papas et~al.(2012)Papas, Houit, Nowrouzezahrai, Gross, and Jarosz]{papas_magic_2012}
Marios Papas, Thomas Houit, Derek Nowrouzezahrai, Markus Gross, and Wojciech Jarosz.
\newblock The magic lens: refractive steganography.
\newblock \emph{ACM Transactions on Graphics}, 31\penalty0 (6):\penalty0 1--10, 2012.

\bibitem[Paszke et~al.(2019)Paszke, Gross, Massa, Lerer, Bradbury, Chanan, Killeen, Lin, Gimelshein, Antiga, Desmaison, Köpf, Yang, DeVito, Raison, Tejani, Chilamkurthy, Steiner, Fang, Bai, and Chintala]{paszke2019pytorch}
Adam Paszke, Sam Gross, Francisco Massa, Adam Lerer, James Bradbury, Gregory Chanan, Trevor Killeen, Zeming Lin, Natalia Gimelshein, Luca Antiga, Alban Desmaison, Andreas Köpf, Edward Yang, Zach DeVito, Martin Raison, Alykhan Tejani, Sasank Chilamkurthy, Benoit Steiner, Lu Fang, Junjie Bai, and Soumith Chintala.
\newblock Pytorch: An imperative style, high-performance deep learning library, 2019.

\bibitem[Perroni-Scharf and Rusinkiewicz(2023)]{perroni-scharf_constructing_2023}
Maxine Perroni-Scharf and Szymon Rusinkiewicz.
\newblock Constructing {Printable} {Surfaces} with {View}-{Dependent} {Appearance}.
\newblock In \emph{{ACM} {SIGGRAPH} 2023 {Conference} {Proceedings}}, pages 1--10, New York, NY, USA, 2023. Association for Computing Machinery.

\bibitem[Podell et~al.(2023)Podell, English, Lacey, Blattmann, Dockhorn, Müller, Penna, and Rombach]{podell2023sdxl}
Dustin Podell, Zion English, Kyle Lacey, Andreas Blattmann, Tim Dockhorn, Jonas Müller, Joe Penna, and Robin Rombach.
\newblock Sdxl: Improving latent diffusion models for high-resolution image synthesis.
\newblock \emph{arXiv}, 2023.

\bibitem[Poole et~al.(2022)Poole, Jain, Barron, and Mildenhall]{Poole2022DreamFusionTU}
Ben Poole, Ajay Jain, Jonathan~T. Barron, and Ben Mildenhall.
\newblock Dreamfusion: Text-to-3d using 2d diffusion.
\newblock \emph{ArXiv}, abs/2209.14988, 2022.

\bibitem[Radford et~al.(2021)Radford, Kim, Hallacy, Ramesh, Goh, Agarwal, Sastry, Askell, Mishkin, Clark, et~al.]{radford2021learning}
Alec Radford, Jong~Wook Kim, Chris Hallacy, Aditya Ramesh, Gabriel Goh, Sandhini Agarwal, Girish Sastry, Amanda Askell, Pamela Mishkin, Jack Clark, et~al.
\newblock Learning transferable visual models from natural language supervision.
\newblock In \emph{International conference on machine learning}, pages 8748--8763. PMLR, 2021.

\bibitem[Ramesh et~al.(2021)Ramesh, Pavlov, Goh, Gray, Voss, Radford, Chen, and Sutskever]{dalle}
Aditya Ramesh, Mikhail Pavlov, Gabriel Goh, Scott Gray, Chelsea Voss, Alec Radford, Mark Chen, and Ilya Sutskever.
\newblock Zero-shot text-to-image generation.
\newblock \emph{ICML}, 2021.

\bibitem[Ramesh et~al.(2022)Ramesh, Dhariwal, Nichol, Chu, and Chen]{dalle2}
Aditya Ramesh, Prafulla Dhariwal, Alex Nichol, Casey Chu, and Mark Chen.
\newblock Hierarchical text-conditional image generation with clip latents, 2022.

\bibitem[Rombach et~al.(2022)Rombach, Blattmann, Lorenz, Esser, and Ommer]{rombach2022high}
Robin Rombach, Andreas Blattmann, Dominik Lorenz, Patrick Esser, and Bj{\"o}rn Ommer.
\newblock High-resolution image synthesis with latent diffusion models.
\newblock In \emph{Proceedings of the IEEE/CVF conference on computer vision and pattern recognition}, pages 10684--10695, 2022.

\bibitem[Saharia et~al.(2021{\natexlab{a}})Saharia, Chan, Chang, Lee, Ho, Salimans, Fleet, and Norouzi]{palette}
Chitwan Saharia, William Chan, Huiwen Chang, Chris~A. Lee, Jonathan Ho, Tim Salimans, David~J. Fleet, and Mohammad Norouzi.
\newblock Palette: Image-to-image diffusion models, 2021{\natexlab{a}}.

\bibitem[Saharia et~al.(2021{\natexlab{b}})Saharia, Ho, Chan, Salimans, Fleet, and Norouzi]{sr3}
Chitwan Saharia, Jonathan Ho, William Chan, Tim Salimans, David~J. Fleet, and Mohammad Norouzi.
\newblock Image super-resolution via iterative refinement, 2021{\natexlab{b}}.

\bibitem[Saharia et~al.(2022)Saharia, Chan, Saxena, Li, Whang, Denton, Ghasemipour, Ayan, Mahdavi, Lopes, Salimans, Ho, Fleet, and Norouzi]{imagen}
Chitwan Saharia, William Chan, Saurabh Saxena, Lala Li, Jay Whang, Emily Denton, Seyed Kamyar~Seyed Ghasemipour, Burcu~Karagol Ayan, S.~Sara Mahdavi, Rapha~Gontijo Lopes, Tim Salimans, Jonathan Ho, David~J Fleet, and Mohammad Norouzi.
\newblock Photorealistic text-to-image diffusion models with deep language understanding.
\newblock \emph{arXiv:2205.11487}, 2022.

\bibitem[Samsudin(2023)]{samsudin_ambigrams_2023}
Noufal Samsudin.
\newblock Generating ambigrams using deep learning: A typography approach, 2023.
\newblock unpublished work.

\bibitem[Sohl-Dickstein et~al.(2015)Sohl-Dickstein, Weiss, Maheswaranathan, and Ganguli]{pmlr-v37-sohl-dickstein15}
Jascha Sohl-Dickstein, Eric Weiss, Niru Maheswaranathan, and Surya Ganguli.
\newblock Deep unsupervised learning using nonequilibrium thermodynamics.
\newblock \emph{ICML}, 2015.

\bibitem[Tancik(2023)]{tancik_illusiondiffusion_2023}
Matthew Tancik.
\newblock Illusion diffusion: optical illusions using stable diffusion, 2023.
\newblock unpublished work.

\bibitem[Tancik et~al.(2020)Tancik, Srinivasan, Mildenhall, Fridovich-Keil, Raghavan, Singhal, Ramamoorthi, Barron, and Ng]{tancik2020fourier}
Matthew Tancik, Pratul~P. Srinivasan, Ben Mildenhall, Sara Fridovich-Keil, Nithin Raghavan, Utkarsh Singhal, Ravi Ramamoorthi, Jonathan~T. Barron, and Ren Ng.
\newblock Fourier features let networks learn high frequency functions in low dimensional domains, 2020.

\bibitem[von Platen et~al.(2022)von Platen, Patil, Lozhkov, Cuenca, Lambert, Rasul, Davaadorj, and Wolf]{diffusers}
Patrick von Platen, Suraj Patil, Anton Lozhkov, Pedro Cuenca, Nathan Lambert, Kashif Rasul, Mishig Davaadorj, and Thomas Wolf.
\newblock Diffusers: State-of-the-art diffusion models.
\newblock \url{https://github.com/huggingface/diffusers}, 2022.

\bibitem[Wittgenstein(1953)]{wittgenstein_philosophical_1953}
Ludwig Wittgenstein.
\newblock \emph{Philosophical investigations.}
\newblock Macmillan, Oxford, England, 1953.
\newblock (Part 2, Section 11).

\bibitem[Yeh et~al.(2023)Yeh, Hsieh, Gao, Yang, Oh, and Gong]{yeh2023navigating}
Shin-Ying Yeh, Yu-Guan Hsieh, Zhidong Gao, Bernard~BW Yang, Giyeong Oh, and Yanmin Gong.
\newblock Navigating text-to-image customization: From lycoris fine-tuning to model evaluation.
\newblock \emph{arXiv preprint arXiv:2309.14859}, 2023.

\bibitem[Yu et~al.(2022)Yu, Xu, Koh, Luong, Baid, Wang, Vasudevan, Ku, Yang, Ayan, Hutchinson, Han, Parekh, Li, Zhang, Baldridge, and Wu]{parti}
Jiahui Yu, Yuanzhong Xu, Jing~Yu Koh, Thang Luong, Gunjan Baid, Zirui Wang, Vijay Vasudevan, Alexander Ku, Yinfei Yang, Burcu~Karagol Ayan, Ben Hutchinson, Wei Han, Zarana Parekh, Xin Li, Han Zhang, Jason Baldridge, and Yonghui Wu.
\newblock Scaling autoregressive models for content-rich text-to-image generation.
\newblock \emph{arXiv:2206.10789}, 2022.

\end{thebibliography}
}

\clearpage
\setcounter{page}{1}

\maketitlesupplementary

\appendix


\begin{figure}
    \centering
    \href{https://diffusionillusions.com}{\includegraphics[width=1\linewidth]{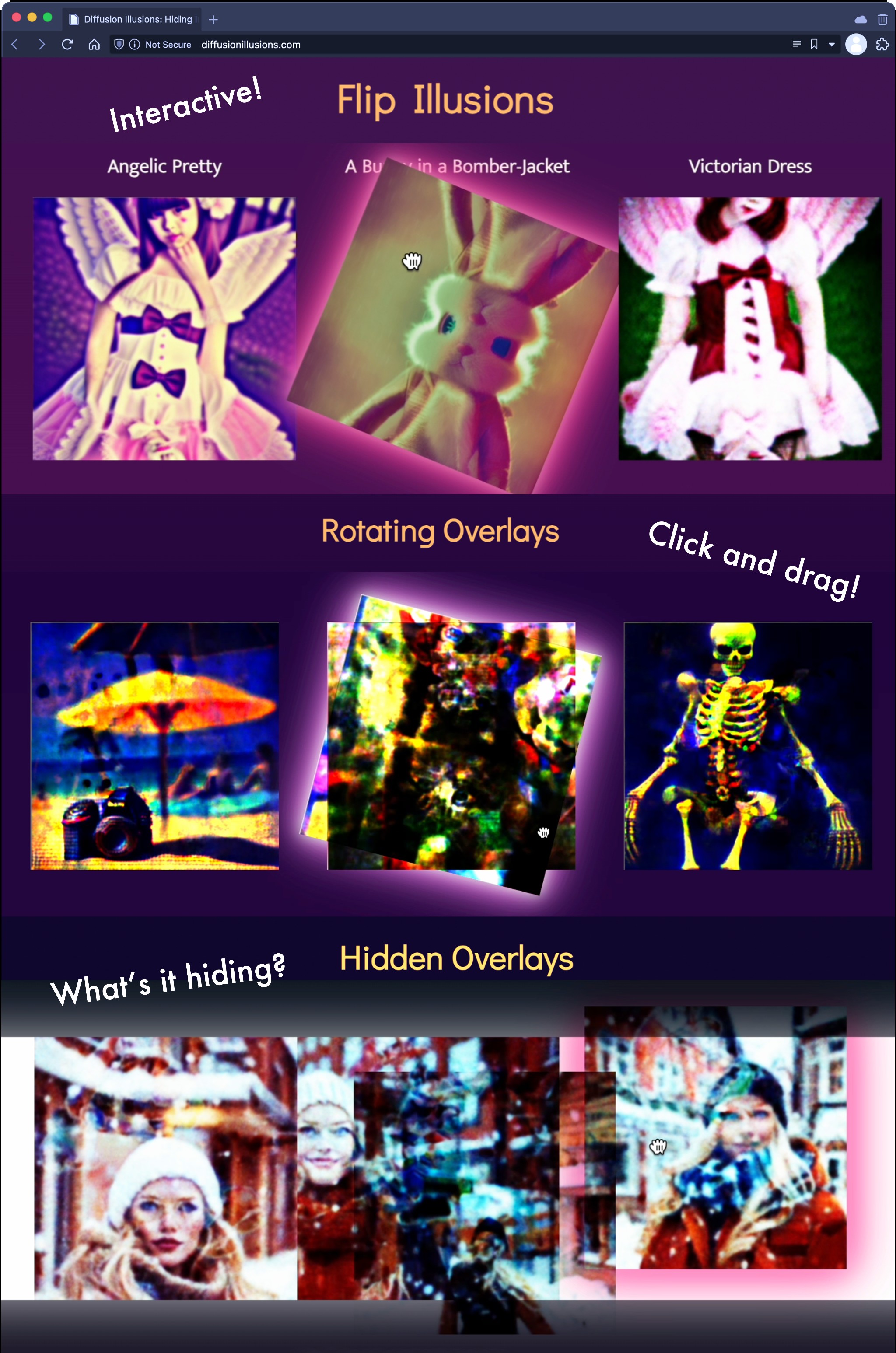}
    }\href{https://diffusionillusions.com}{https://diffusionillusions.com} \caption{Please visit our project page --- it contains fully interactive simulations of all illusions in this paper, as well as many more!}
    \label{fig:enter-label}
\end{figure}

\section{Implementation Details}

\subsection{Brightness Constant}
In the actual implementation, you'll see we multiply our derived overlay images by a scalar ``brightness constant'' $k$, that is chosen based on the type of illusion. This constant is visible in the given pseudocode --- please see how it is used there. This is because in real life, when viewing the hidden overlay and rotating overlay illusions, the backlight can be arbitrarily bright. Without this term, the derived images obtained from overlaying other images would necessarily be darker than their prime images, because images have values between 0 and 1, and the product between any two numbers between 0 and 1 are guaranteed to be 1 or less. 

Because the hidden character illusion deals with 4 overlays, it benefits from a higher brightness constant than the rotation overlay illusion ($k=3$ vs $k=2$). The brightness constant $k$ is not applicable for the flip illusion, as it does not deal with overlay transparencies.

\subsection{Static Targets}
When creating an illusion, usually text prompts are used for all values of $T$. However, it is possible to specify a fixed image target by setting $T$ as an image instead. This allows us to hide specific images such as QR codes, nyan cat, pentagrams, or even entire segments of text (see \cref{fig:galhid}). Instead of applying score distillation loss for example, we regress torwards that given image. Please see the below pseudocode for an exact implementation.

\subsection{Libraries}
We use SDXL as our latent diffusion model \cite{podell2023sdxl}. Our SDEdit implementation of SDXL comes from \cite{diffusers}, using PyTorch \cite{paszke2019pytorch}. Our implementation of fourier feature networks is directly adapted from the TRITON \cite{Burgert2022}, using the default parameters for their Neural Neural Textures. Our implementation of Score Distillation Loss comes from Peekaboo \cite{Burgert2022PeekabooTT}.

\subsection{Pseudocode}
\label{sec:pseudocode}
In this subsection, we show a Python-like pseudocode that outlines the exact process of creating the algorithm. Refer to the next page.

\begin{lstlisting}
#           _  ___  _  _      __ ___  _
#          | \  |  |_ |_ | | (_   |  / \ |\ |
#          |_/ _|_ |  |  |_| __) _|_ \_/ | \|
#          ___            __ ___  _        __
#           |  |  |  | | (_   |  / \ |\ | (_
#          _|_ |_ |_ |_| __) _|_ \_/ | \| __)
#       
#       
#                 _   __  _      _   _
#                |_) (_  |_ | | | \ / \
#                |   __) |_ |_| |_/ \_/
#                     _  _   _   _
#                    /  / \ | \ |_
#                    \_ \_/ |_/ |_


###### PART 1: Initialization
if ILLUSION_TYPE=='FLIP':
    n = 1 #Number of Prime images
    m = 2 #Number of Derived images
    A = [
        #A stands for Arrangements
        lambda P: P[0],
        lambda P: P[0].rot180(),
    ]
    W = [1, 1] # Importance of each derived image
    T = ['Dog', 'Sloth']

if ILLUSION_TYPE=='ROTATE':
    n = 2 #Two Prime Images: Base, Rotator
    m = 4 #Four Derived Images
    k = 2 #The backlight brightness constant
    A = [
        lambda P: k*P[0]*P[1],
        lambda P: k*P[0]*P[1].rot90(),
        lambda P: k*P[0]*P[1].rot180(),
        lambda P: k*P[0]*P[1].rot270(),
    ]
    W = [1, 1, 1, 1]
    T = ['Dog', 'Cat', 'Man', 'Woman']

if ILLUSION_TYPE=='HIDDEN':
    n = 4 #Two Prime Images: A, B, C, D
    m = 5 #Four Derived Images:
          # A, B, C, D, Hidden
    k = 3 #The backlight brightness constant
    A = [
        lambda P: P[0],
        lambda P: P[1],
        lambda P: P[2],
        lambda P: P[3],
        lambda P: k*P[0]*P[1]*P[2]*P[3],
    ]
    W = [1, 1, 1, 1, 3] # Prioritize the hidden image
    T = ['Dog', 'Penguin', 'Giraffe', 'Cow', 'Cat']
    ## OR, to use a QR code or another specific image...
    T = ['Dog', 'Penguin', 'Giraffe', 'Cow',
         load_image('qr_code.png') ]

assert len(T) == len(A) == len(W) == m

# Initialize all prime images
P = [RgbFourierFeatureNetwork(resolution=(512,512))
     for _ in range(n)]

# Initialize our latent diffusion model
F = StableDiffusion()

# We optimize the prime images via gradient descent. 
optim = SGD(P.parameters())


###### PART 2: Helper Functions
def score_distill_loss(image, prompt):
    #Same loss proposed in DreamFusion -
    # but with a latent diffusion model
    image_latent = F.encode_image(image)
    timestep = random_int(0, F.max_timestep)
    noise = F.get_noise(timestep)
    noised_latent = F.add_noise(
        image_latent, noise, timestep
    )    
    with torch.no_grad():
        text_embed = F.clip.embed(prompt)        
        pred_noise = F.unet(
            noised_latent, text_embed, timestep
        )    
    return abs(noise - pred_noise).sum()

def image_similarity(a, b):
    #Our image similarity metric 
    return SSIM(a,b) - MSE(a,b)

def img2img(image, prompt, strength):
    #Based on SDEdit - simplified here
    #When strength=1, the entire image is replaced
    #When strength=0, nothing is changed
    image_latent = F.encode_image(image)
    timestep = int(strength * F.max_timestep)
    noise = F.get_noise(timestep)
    noised_latent = F.add_noise(
        image_latent, noise, timestep
    )

    #Perform diffusion as normal, but starting from
    #our noised_latent instead of pure noise
    diffused_latent = F.text_to_image(
        prompt,
        initial_latent=noised_latent,
        initial_timestep=timestep,
    )

    new_image = F.decode_image(diffused_latent)
    return new_image


###### PART 3: Optimization

#Phase 1: Score Distillation Loss

for iteration in range(10000):
    loss = 0
    for a,t,w in zip(A,T,W):
        # Derived image d
        # comes from an arrangement of prime images
        d = a(P)
        if isinstance(t, str):
            loss += w * score_distill_loss(d, t)
        elif is_image(t):
            # For hiding custom images such as QR codes
            loss -= w * image_similarity(d, t)
    optim.update(loss) # Take a gradient descent step

#Phase 2: Dream-Target Loss

#Start from strength = .90 instead of 1
# in order to use the results from Phase 1
schedule = [.90, .89, .88 ... .03, .02, .01]

for strength in schedule:
    # Define the image translation function
    G = lambda text,image: img2img(text,image,strength)

    # Step 1: Set our Dream-Targets
    Z = []
    for a, t in zip(A,T):
        if isinstance(t, str):
            # Tweak a derived image to get a new target
            d = a(P)
            z = G(t, d)
        elif is_image(t):
            #Use a predefined target (e.g. a QR code)
            z = t
        Z.append(z)

    # Step 2: Approach our Dream-Targets
    for iteration in range(1000):
        #Optimize P so that D approaches T

        loss = 0
        for a,z,w in zip(A,Z,W):
            d = a(P)
            loss -= w * image_similarity(d, z)
            
        # Take a gradient descent step
        optim.update(loss)

        
###### PART 4: Fabrication
        
#We're done! Return the primes - 
# and print them out physically! 
printed_P = send_to_laser_printer(P)

#Oh, and also, make sure someone uses them...
fun = have_human_arrange_the_illusions(printed_P)
\end{lstlisting}

\section{Extended Quantitative Evaluation}
\label{sec:more_eval}
\subsection{Quantitative Evaluation Details}
This section provides more details and additional experiments regarding benchmarking the derived images of Hidden Overlay Illusion and Rotation Overlay Illusion.

\noindent\textbf{Textual Prompts} \quad The set of image styles $T^s$ is listed as follow where \texttt{<s>} stands for the \textit{subject} token:

\begin{quote}
Style 1: \textit{3d pixar style render animation of a \texttt{<s>}}

Style 2: \textit{an award winning photograph of a \texttt{<s>}}

Style 3: \textit{an award winning photograph of a \texttt{<s>} in the deep jungle}

Style 4: \textit{an award winning photograph of a \texttt{<s>} in times square}
\end{quote}

The subject set $T^o$ contains subjects from PASCAL VOC dataset~\cite{everingham2010pascal}: \textit{aeroplane}, \textit{bicycle}, \textit{bird}, \textit{boat}, \textit{bottle}, \textit{bus}, \textit{car}, \textit{cat}, \textit{chair}, \textit{cow}, \textit{dining table}, \textit{dog}, \textit{horse}, \textit{motorbike}, \textit{potted plant}, \textit{sheep}, \textit{sofa}, \textit{train}, \textit{tv/monitor}.

\noindent\textbf{Additional Evaluation Metrics} \quad We further extend the evaluation introduced in the main paper by including more metrics in each aspect:

\begin{itemize}
    \item \textit{Controllability} We take advantage of a vision language model (VLM) LLaVA-1.5~\cite{liu2023visual, liu2023improved} to measure the similarity between the image and the textual prompt. The instruction sent to the VLM is 
    \begin{quote}
    \textit{Give a single score from 0 to 10 regarding how well the image looks like a \texttt{<s>}. A higher score means the image generally looks similar to a \texttt{<s>}. Only return the score.}
    \end{quote} 
    where \texttt{<s>} stands for the \textit{subject} token and it will substituted by the actual subject for a specific image.
    \item \textit{Diversity} Recent research~\cite{darcet2023vision} suggests that the feature from the original DINOv2 might suffer from abnormal patches corresponding to the plain areas of the image. Therefore, we report a new \textit{Vendi Score} using the feature from  DINOv2+reg~\cite{darcet2023vision}. 
    \item \textit{Aesthetics} Similar to Controllability, we collect an aesthetics score from LLaVA-1.5 using the following instruction:
    \begin{quote}
    \textit{Give a single score from 0 to 10 regarding how well this image looks. A higher score means the image generally looks more natural and has fewer artifacts. Only return the score.}
    \end{quote} 
\end{itemize}

In all metrics, the vision encoder of CLIP and the backbone of all DINO variants is a ViT-L/14~\cite{dosovitskiy2020image}. The version of LLaVA-1.5 we utilized is fine-tuned from Vicuna-13B.

\begin{figure}[h]
  \centering
  \includegraphics[width=1\linewidth]{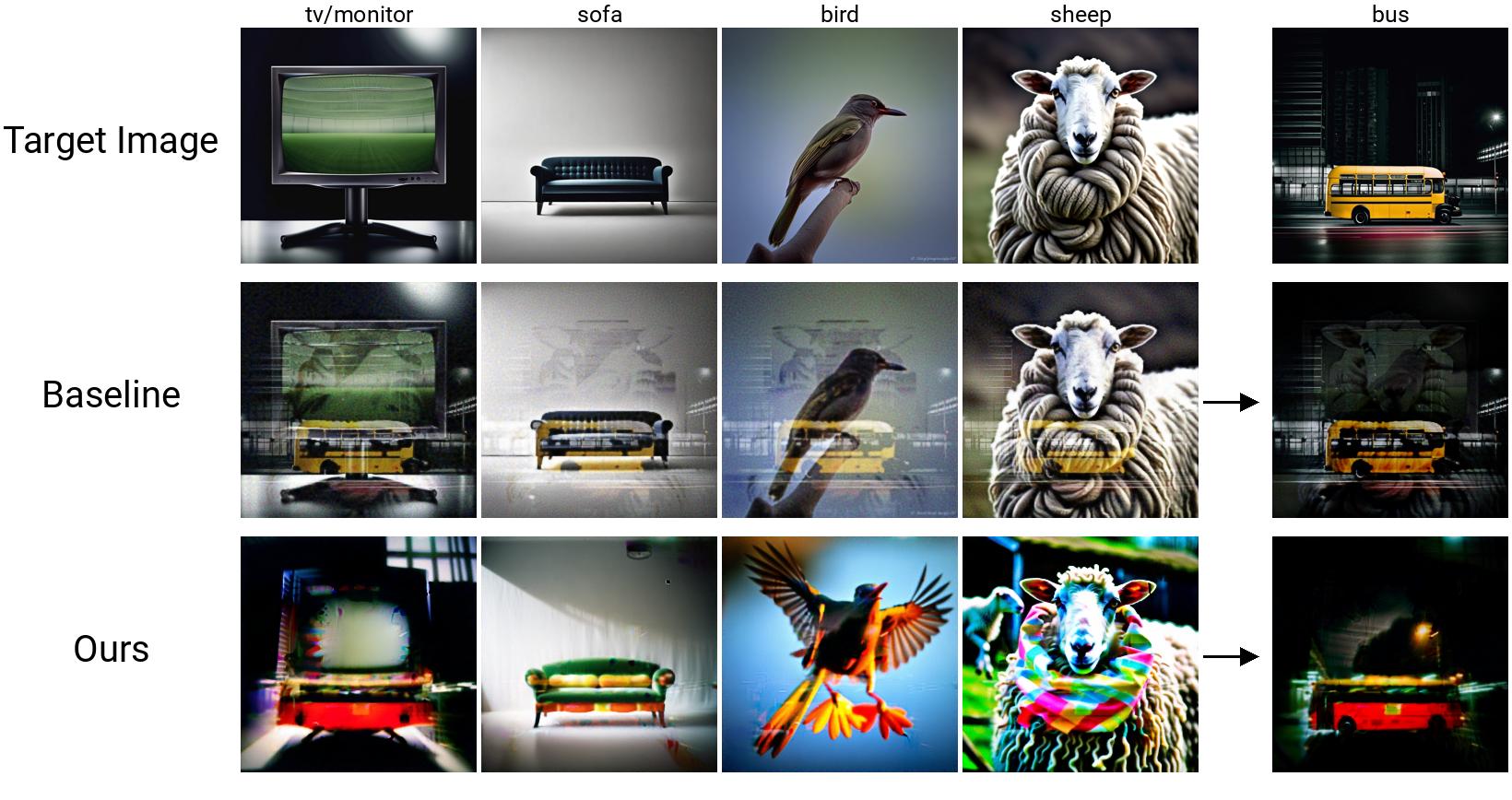}
  \caption{Examples of our method and the baseline, starting from the same target image. Note how in the baseline, you can see the sheep in the bus image and the bus in the sheep image - which is why its independence score is lower.}
  \label{fig:baseline-demo}
\end{figure}

\subsection{Extended Results of Hidden Overlay Illusion}
\cref{fig:baseline-demo} presents comparative examples between the proposed method and the established baseline, starting from the same target image. The images from the baseline are heavily interfered with by others in the same group and the overlay image. 

\begin{figure}[h]
  \centering
  \includegraphics[width=1\linewidth]{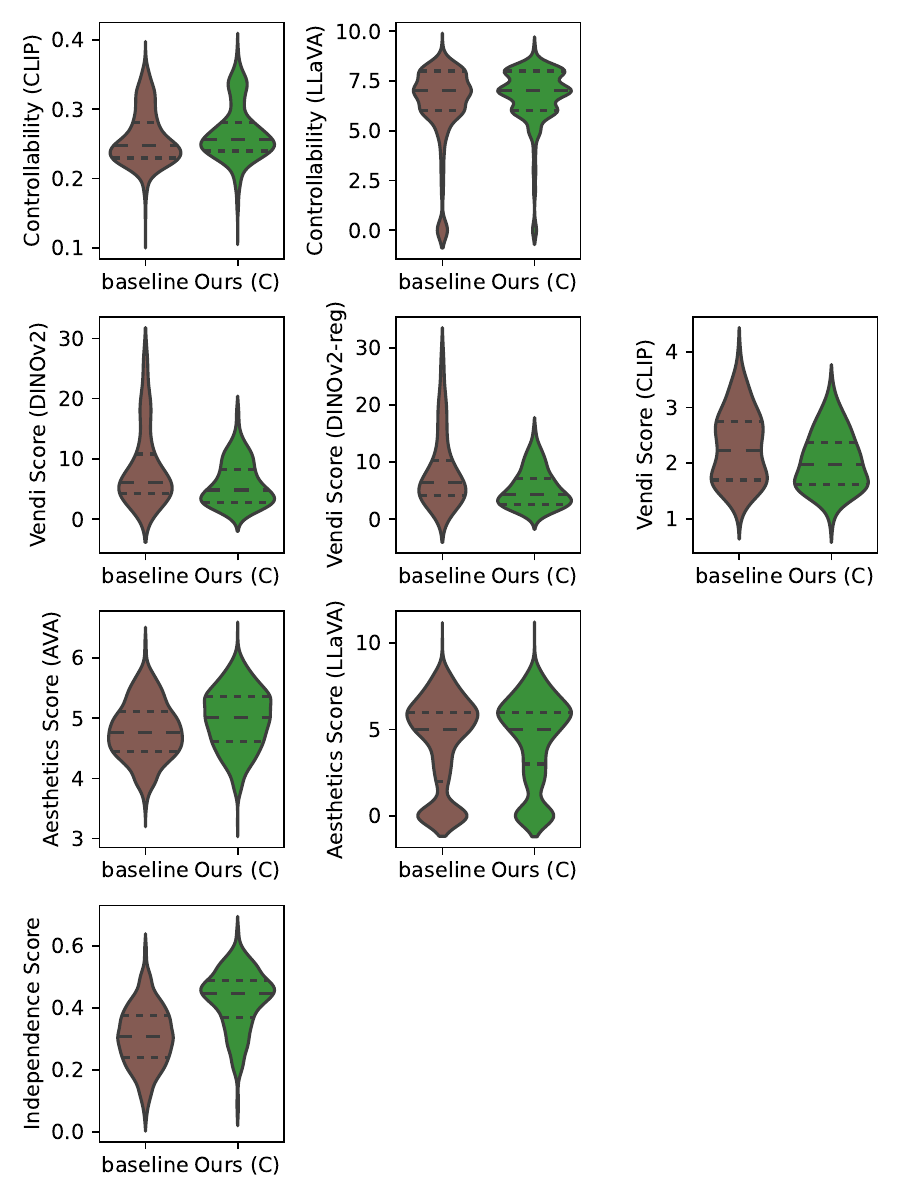}
  \caption{Full evaluation on Hidden Overlay Illusion, each row is a group of thematically-aligned figures.}
  \label{fig:big-baseline-fig-full}
\end{figure}

\begin{figure}[h]
  \centering
  \includegraphics[width=0.49\textwidth]{figs/final-clip_score.png}
  \includegraphics[width=0.49\textwidth]{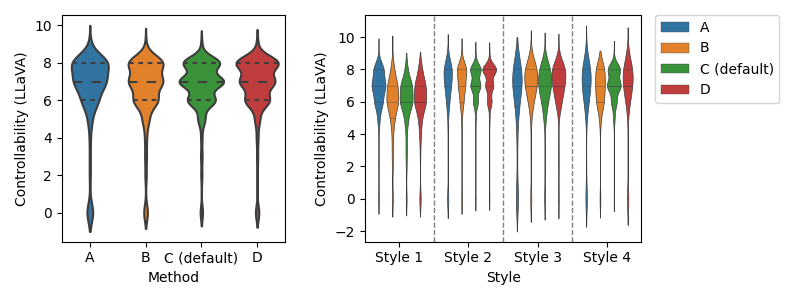}
  \caption{Controllability score distributions over methods (left) and styles (right). A, B, C, D stands for four variants of our method}
  \label{fig:big-score-fig-full1}
\end{figure}

\begin{figure}[h]
  \centering
  \includegraphics[width=0.49\textwidth]{figs/final-vendi_score_dino.png}
  \includegraphics[width=0.49\textwidth]{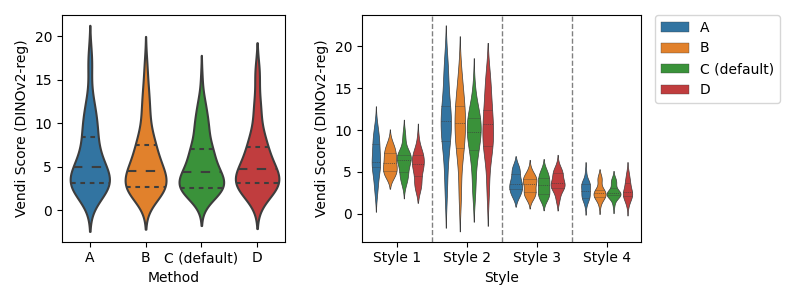}
  \includegraphics[width=0.49\textwidth]{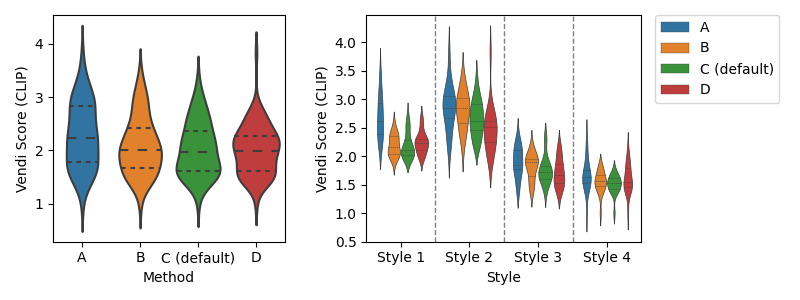}
  \caption{Diversity score distributions over methods (left) and styles (right). A, B, C, D stands for four variants of our method}
  \label{fig:big-score-fig-full2}
\end{figure}

\begin{figure}[h]
  \centering
  \includegraphics[width=0.49\textwidth]{figs/final-aesthetics_score.png}
  \includegraphics[width=0.49\textwidth]{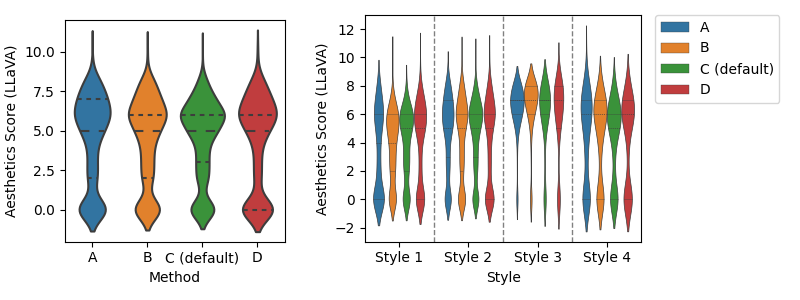}
  \caption{Aesthetics score distributions over methods (left) and styles (right). A, B, C, D stands for four variants of our method}
  \label{fig:big-score-fig-full3}
\end{figure}

\cref{fig:big-baseline-fig-full}, \cref{fig:big-score-fig-full1}, \cref{fig:big-score-fig-full2} and \cref{fig:big-score-fig-full3} show full evaluation results of the derived images from baseline and four variants of our method. 
The advantages of our method compared to the baseline are further supported by the new metrics introduced in this section, like better Controllability and Aesthetics Score from LLaVA (see \cref{fig:big-baseline-fig-full}). 
Meanwhile, LLaVA has relatively less bias on art styles and different subjects (\cref{fig:big-score-fig-full1} and \cref{fig:big-score-fig-full3})

\subsection{Results of Rotation Overlay Illusion}
\begin{figure}
  \centering
  \includegraphics[width=1\linewidth]{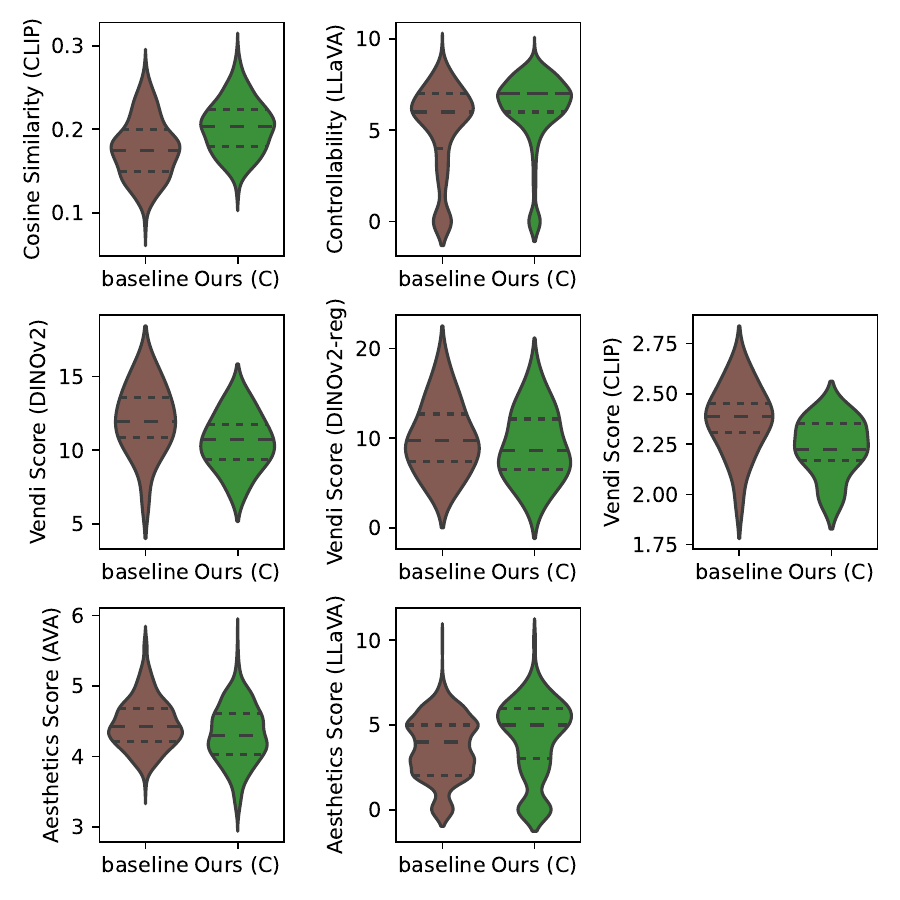}
  \caption{Full evaluation on Rotation Overlay Illusion, each row is a group of thematically-aligned figures.}
  \label{fig:big-baseline-fig-rot}
\end{figure}

We further benchmark the performance of Rotation Overlay Illusion. 
The evaluation follows the same protocol as the Hidden Overlay Illusion except that each group of Rotation Overlay Illusion images only has $4$ derived images, which require $4$ textual prompts at a time and we focus on one style:
\begin{quote}
\textit{a beautiful award-winning royalty-free full-frame stock photo of an isolated \texttt{<s>}}.
\end{quote}
The result is presented in \cref{fig:big-baseline-fig-rot}. Our method is significantly better than the baseline in terms of controllability (CLIP cosine similarity) and Aesthetics Score. 

\begin{figure}
  \centering
  \includegraphics[width=1\linewidth]{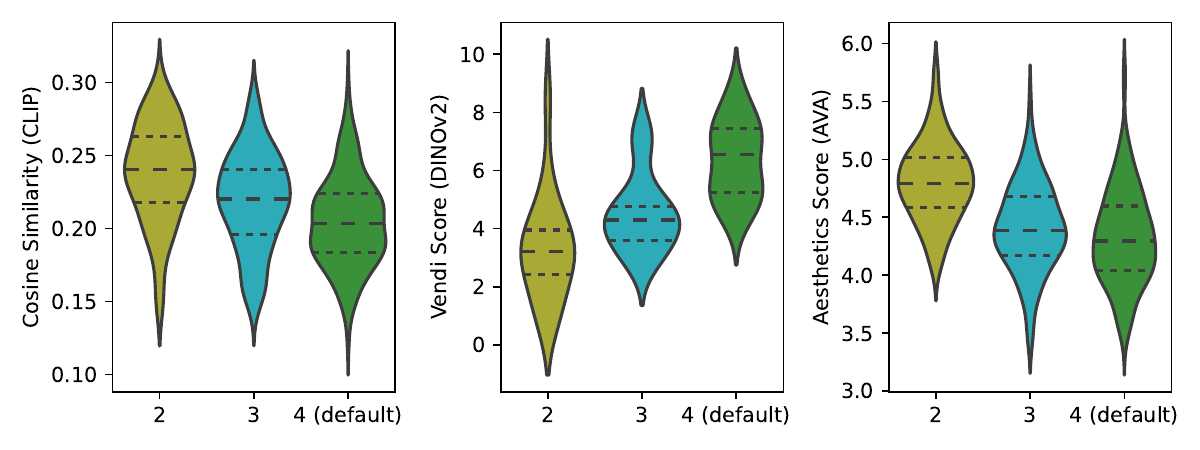}
  \caption{Ablation on the number of the derived images in Rotation Overlay Illusion}
  \label{fig:big-baseline-fig-rot-task}
\end{figure}

\begin{figure}
  \centering
  \includegraphics[width=1\linewidth, trim={0, 0, 3cm, 0}, clip]{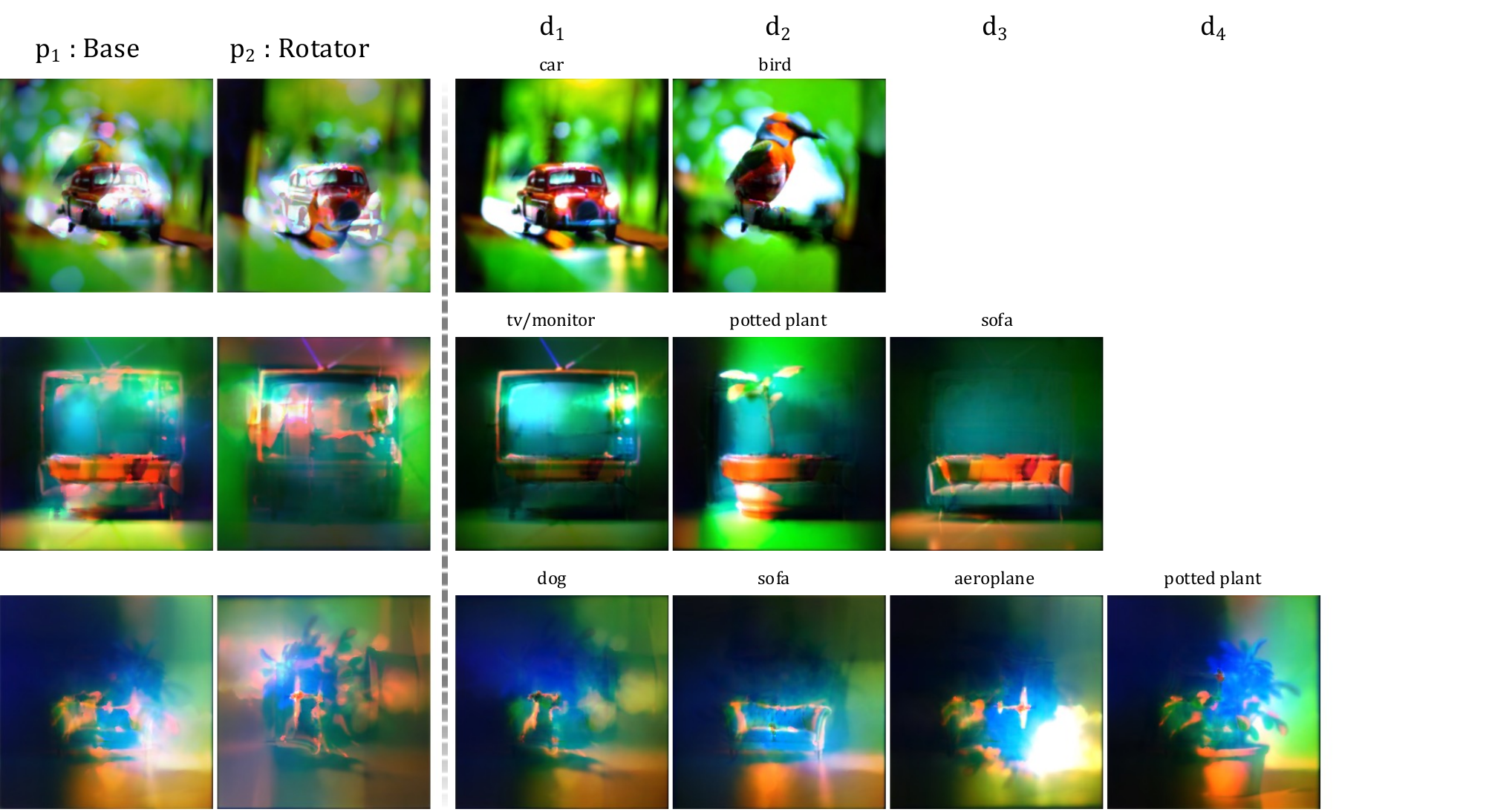}
  \caption{Qualitative results of Rotation Overlay Illusions with different numbers of derived images}
  \label{fig:demo-rot-task}
\end{figure}

\noindent\textbf{Ablation on the Number of Derived Images} \quad In this paper, by default we discuss a challenging rotation overlay illusion task where two prime images need to `encode' four derived images. In this section, we conduct an ablation on the number of derived images. We perform an ablation study on the number of derived images, specifically focusing on cases with 2 to 4 derived images. 
Our hypothesis posits that reducing the number of derived images eases generation constraints, potentially enhancing image quality. 
This is corroborated by \cref{fig:big-baseline-fig-rot-task}, which demonstrates improved image-text alignment and aesthetic scores in simpler tasks. 
Conversely, we observe a divergent trend in diversity, suggesting the interference between multiple derived images.
\cref{fig:demo-rot-task} presents a qualitative comparison between problem formulations.

\section{Fabrication Details}
All of the illusions we present are realizable in the real world in physical form. To create a flip illusion in real life is quite easy - just print out one of the images onto a sheet of paper using a regular color laser printer. 

The hidden overlay and rotation overlay can also be created with a basic color laser printer, and this is how we made all of the photographic examples in this paper. Searching ``transparency film'' online will yield many cheap transparent plastic films that are laser-printer compatible (a pack of 100 sheets sells for about \$20 USD). However, after printing onto these overlays, it is useful to laminate them, as the ink can be easily scratched off. We do this with a basic thermal lamination machine that can also be purchased cheaply online.

After they have been printed, laminated, and cut appropriately - place the stacked transparencies over a source of light. We use a backlight extracted from an old LCD monitor for our photos in this paper. However, any backlight will do - holding them up to a bright window with outdoor sunlight works quite well too! 

Since we model the light filtering process as multiplication, and multiplication is commutative, our modeling process assumes that the ordering of the layers doesn't matter. This is true in real life as well - with sufficient backlighting, you will get the same visual result whether transparency $p_1$ is on the top or on the bottom. However in practice, since some light reflects off the top transparency, it won't be perfectly identical.

Additionally, we found that inserting a thin layer of water between the transparent overlay sheets further enhances the visual effect, and slightly reduces the need for as strong of a backlight. We suspect this is because it eliminates the air gap between the sheets, leading to a smaller difference in the index of refraction. This is not necessary, but can somewhat enhance the clarity of the illusion.

\begin{figure*}[t]
    \centering
    \includegraphics[width=.8\linewidth]{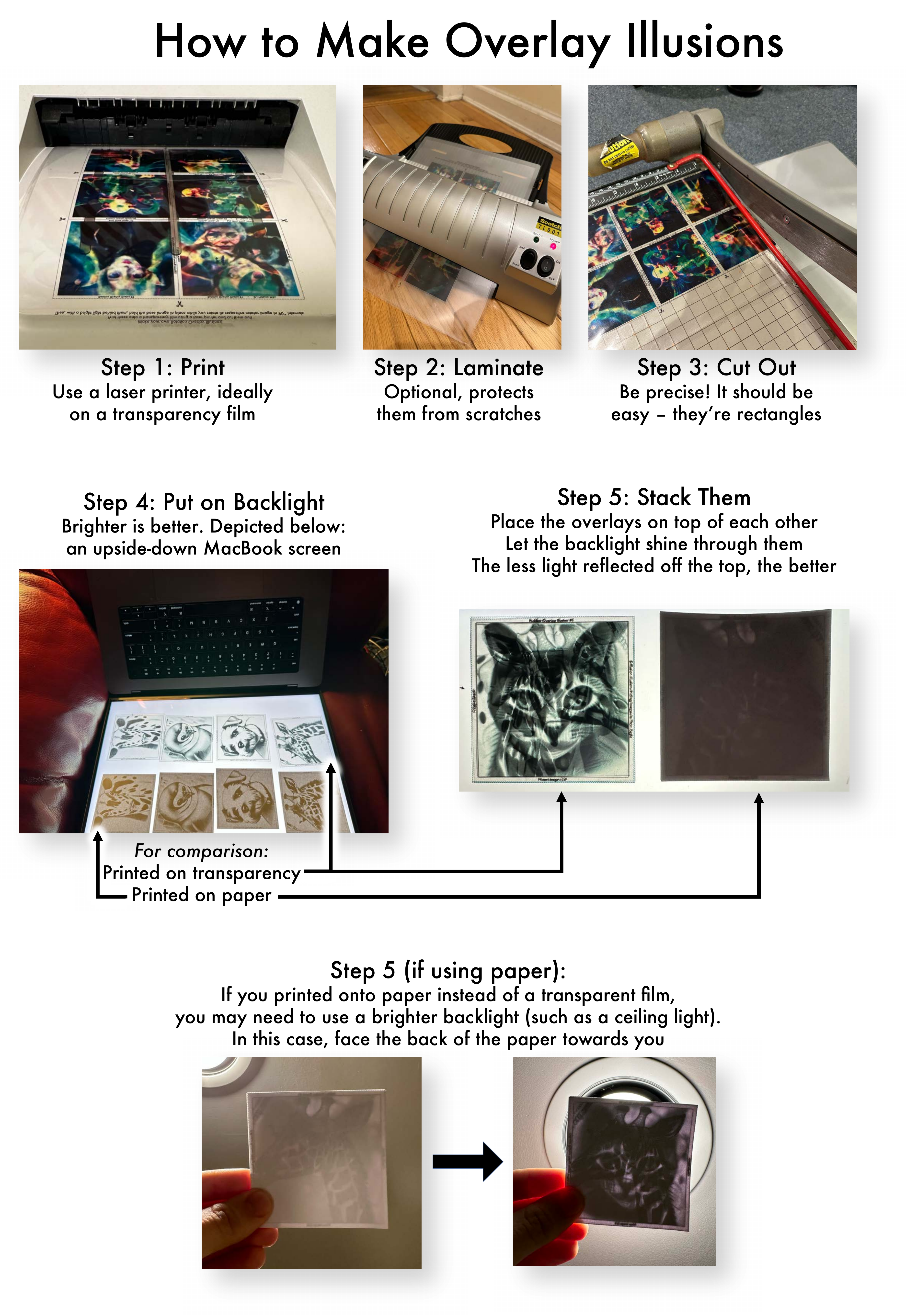}
    \caption{The overlay illusion fabrication process is depicted here. You can use paper instead of transparent films if you wish! But please note that if you do, you may need to use a stronger backlight. The accuracy of your printer can affect the quality of the illusions as well - if the illusion doesn't work, check your toner levels.}
    \label{fig:paperVsPlastic}
\end{figure*}

We would like to point out, however: \textit{you do not strictly need to use transparencies} to create overlay illusions! Regular paper can also work, provided you use a strong enough backlight and use a sufficient amount of ink. We've included a comparison in figure \cref{fig:paperVsPlastic}.

\begin{figure*}
  \centering
  \includegraphics[width=0.35\linewidth]{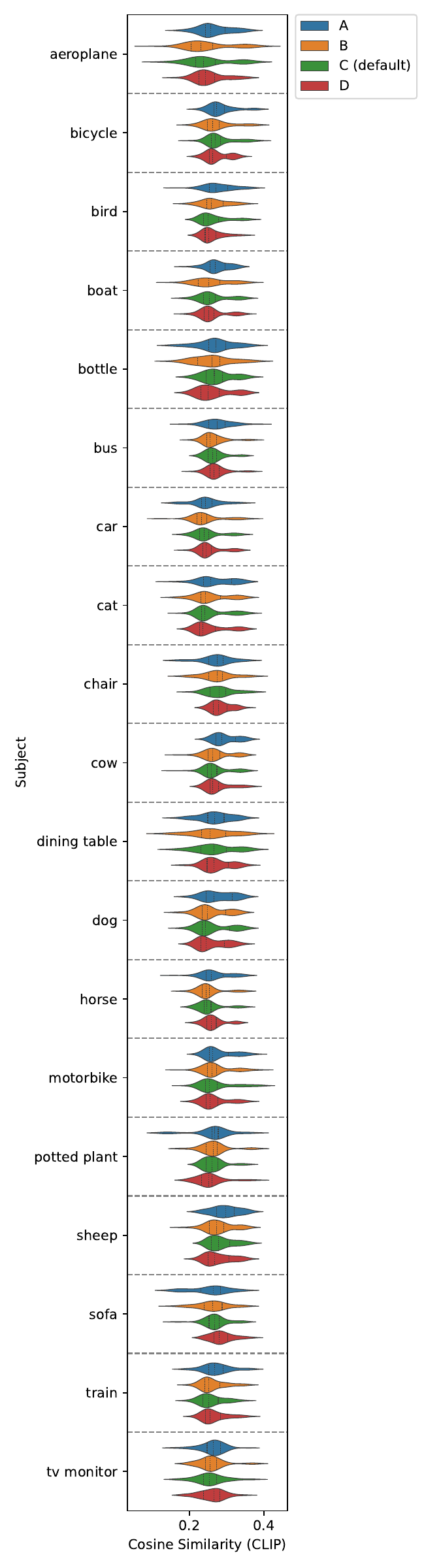}
  \includegraphics[width=0.35\linewidth]{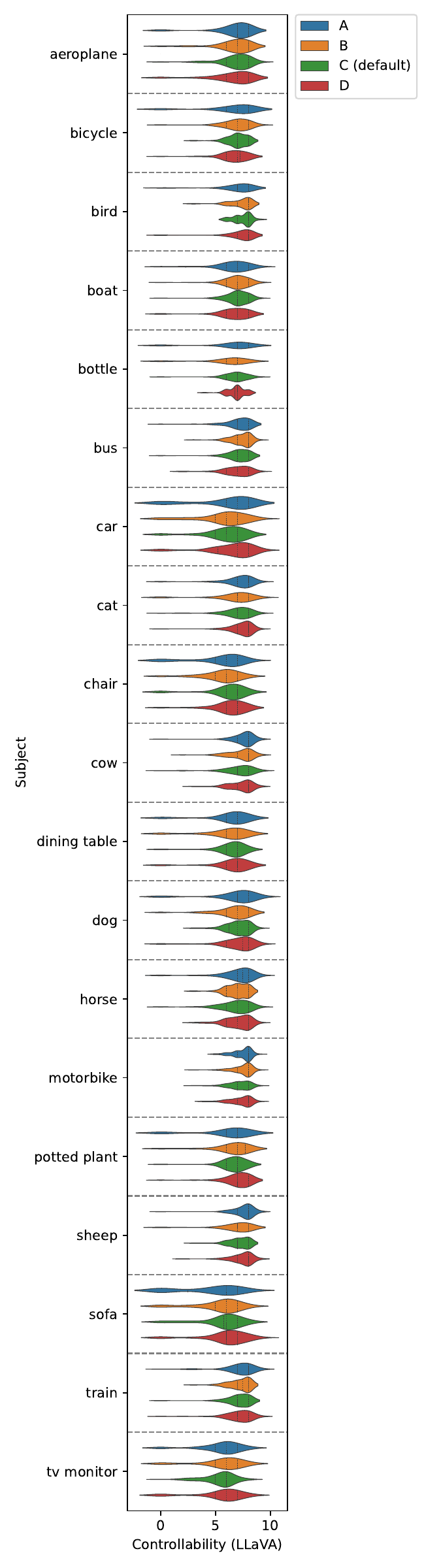}
  \caption{Controllability of Hidden Overlay Illusion over different subjects.}
  \label{fig:score-sub1}
\end{figure*}

\begin{figure*}
  \centering
  \includegraphics[width=0.33\linewidth]{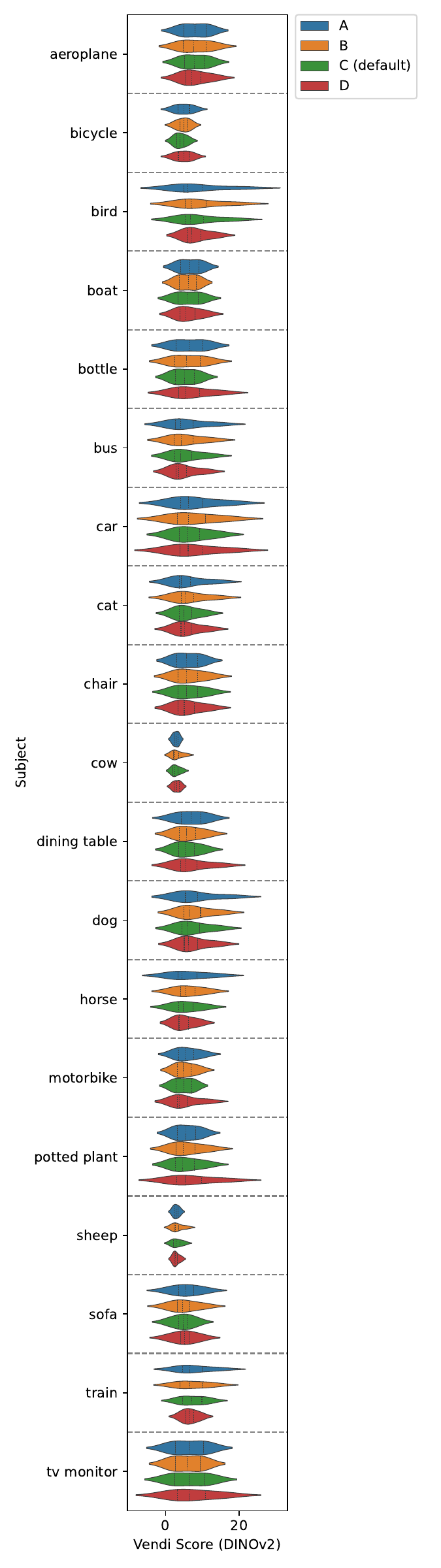}
  \includegraphics[width=0.33\linewidth]{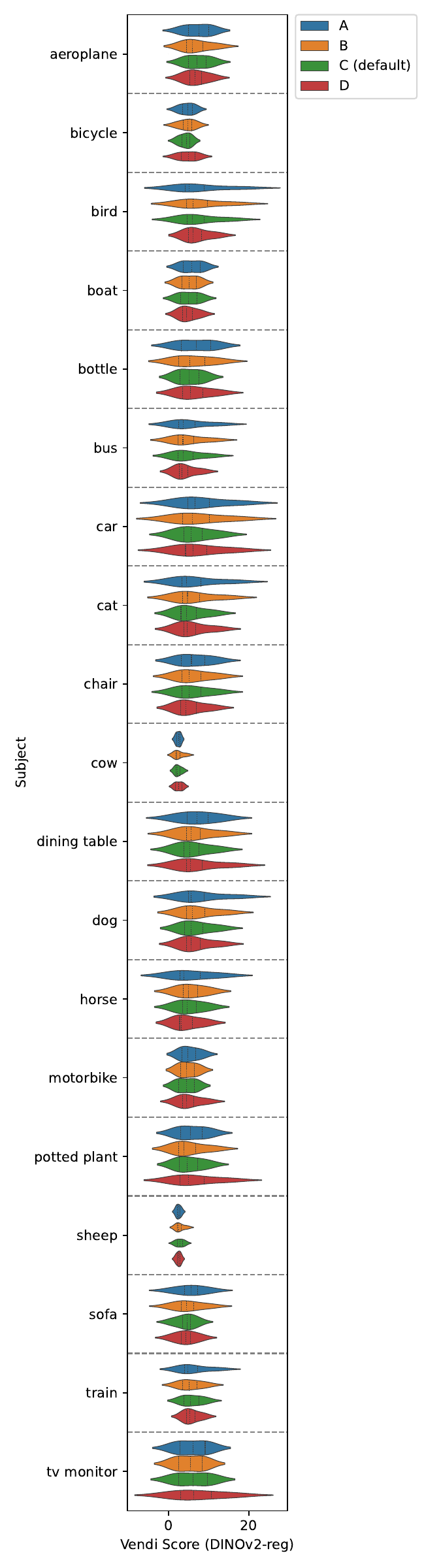}
  \includegraphics[width=0.33\linewidth]{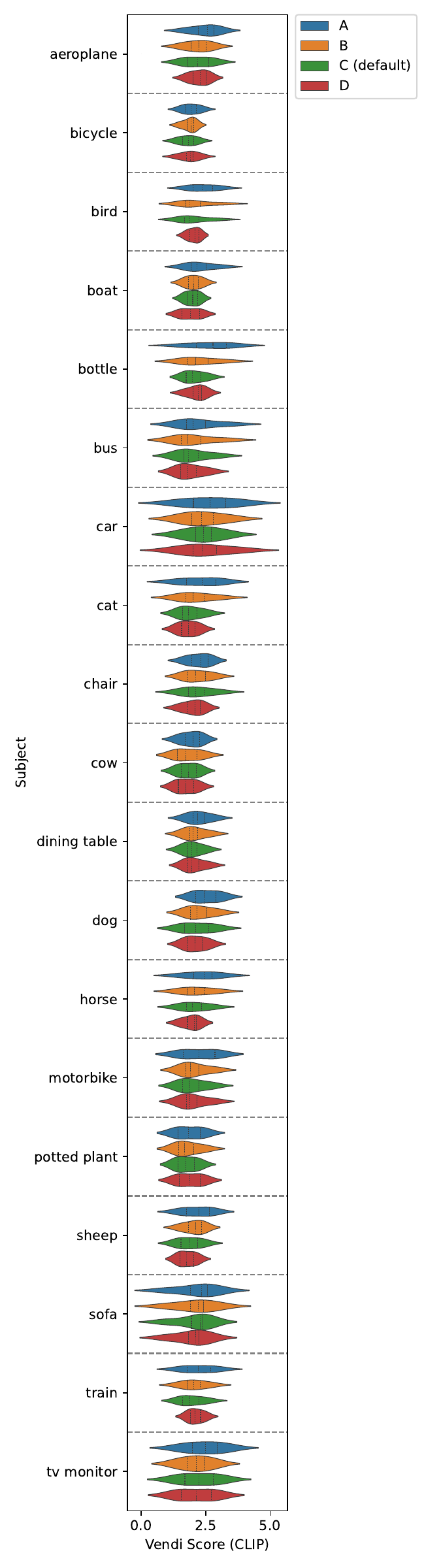}
  \caption{Diversity of Hidden Overlay Illusion over different subjects.}
  \label{fig:score-sub2}
\end{figure*}


\begin{figure*}
\begin{minipage}{0.66\textwidth}
      \centering
      \includegraphics[width=0.49\linewidth]{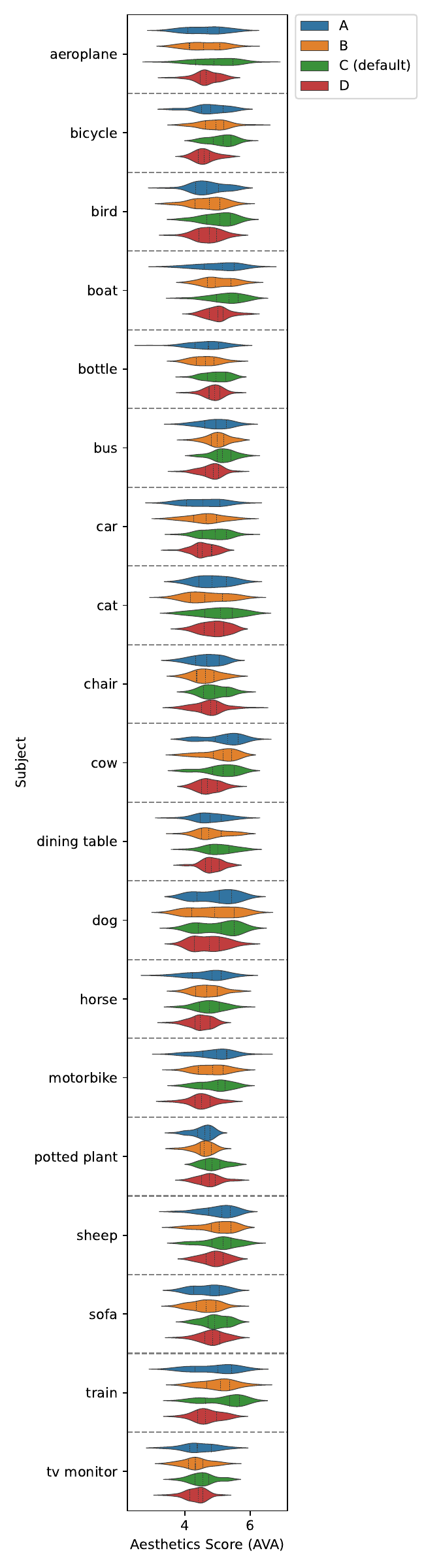}
      \includegraphics[width=0.49\linewidth]{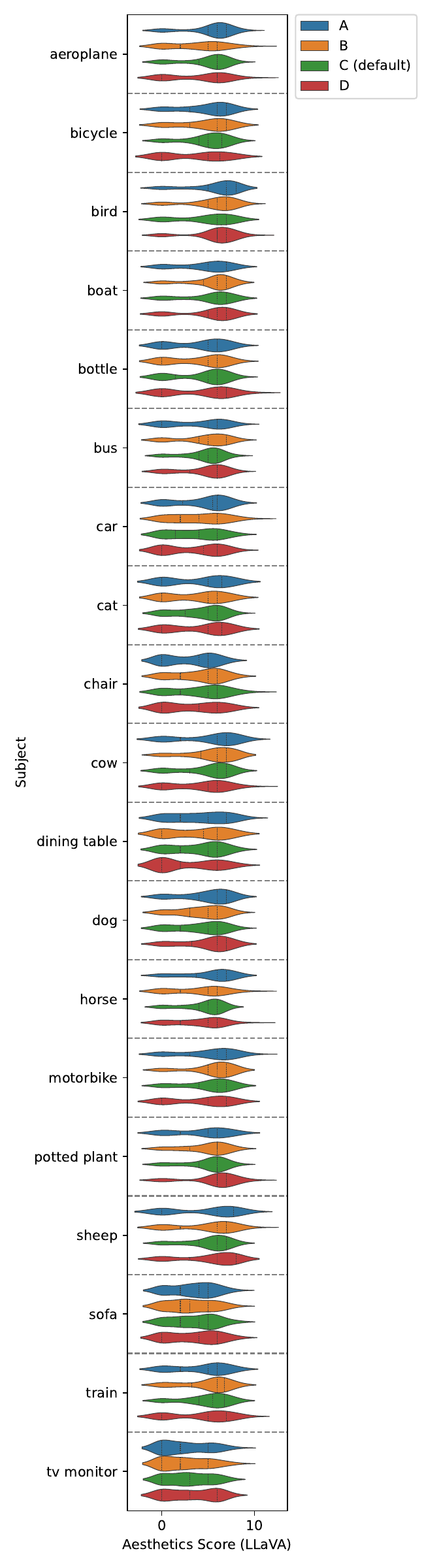}
      \caption{Aesthetics of Hidden Overlay Illusion over different subjects.}
      \label{fig:score-sub3}
\end{minipage}
\begin{minipage}{0.33\textwidth}
      \centering
      \includegraphics[width=0.96\linewidth]{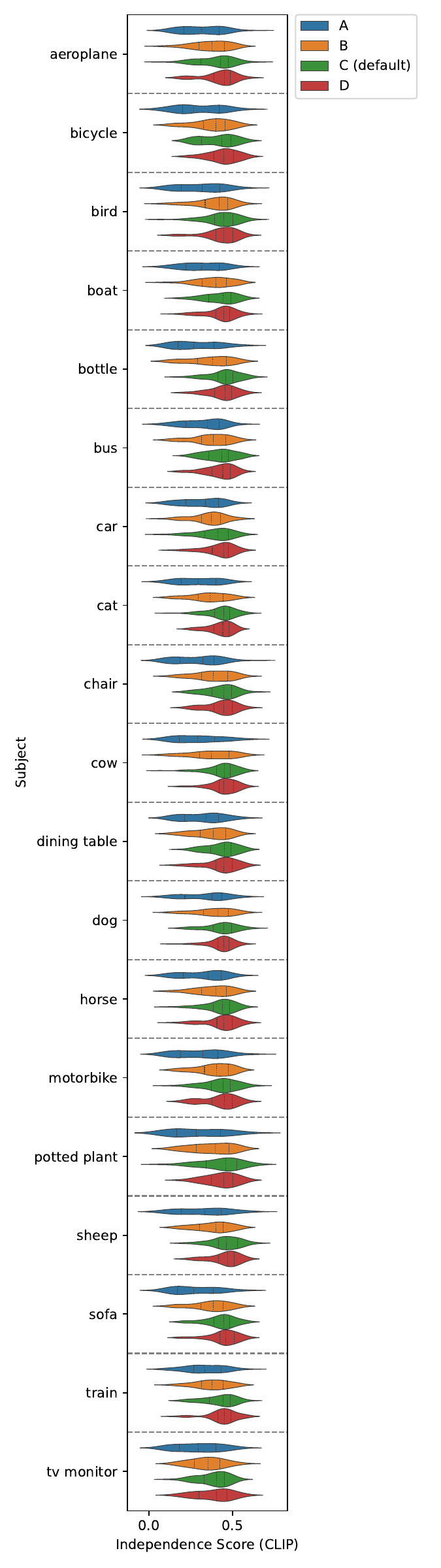}
      \caption{Independence Score of Hidden Overlay Illusion over different subjects.}
      \label{fig:score-sub4}
\end{minipage}
\end{figure*}

\newpage

\begin{figure*}[t]
    \centering
    \includegraphics[width=.88\linewidth]{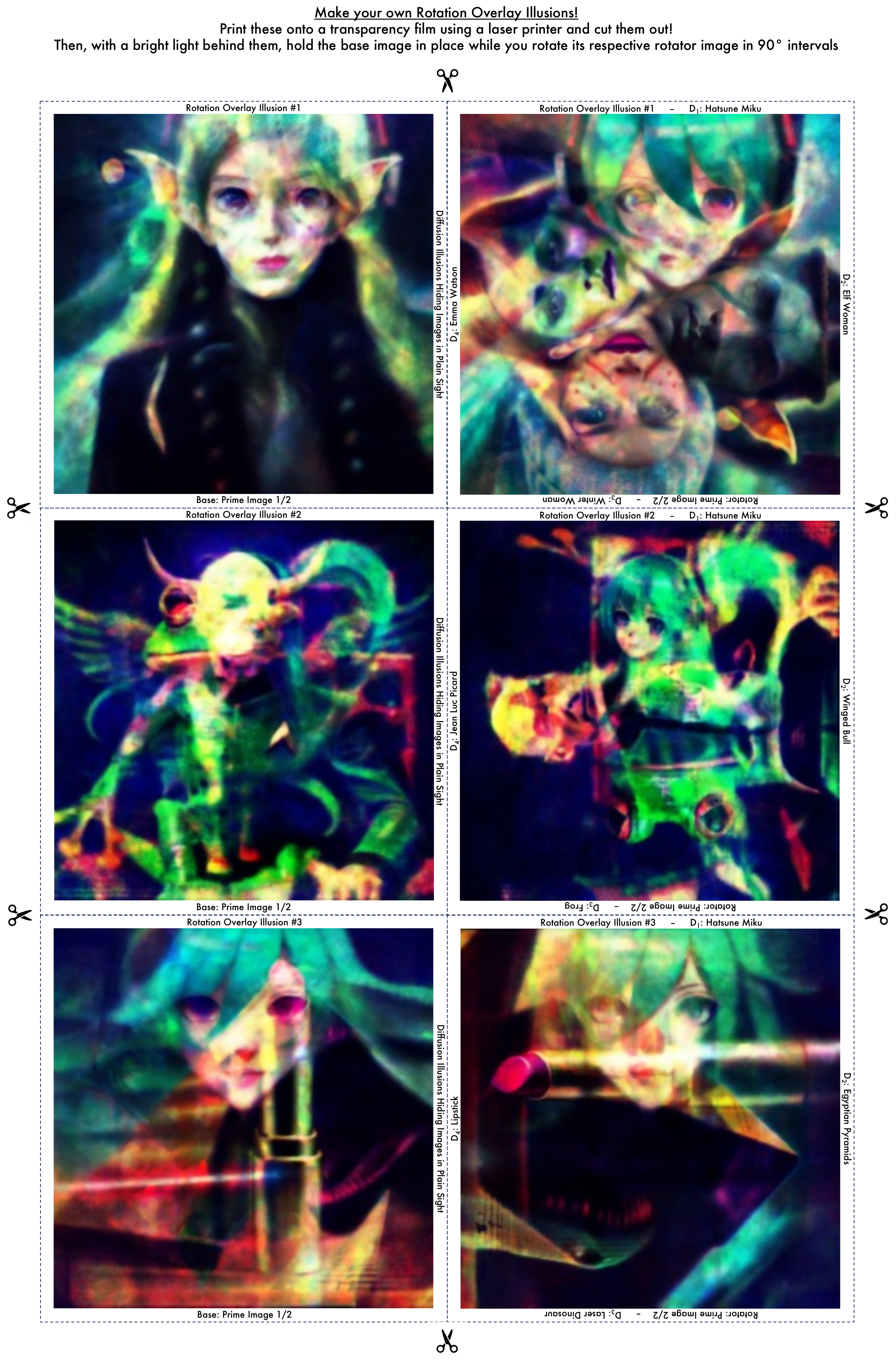}
    \label{fig:make-your-own-rotators}
\end{figure*}

\begin{figure*}[t]
    \centering
    \includegraphics[width=\linewidth]{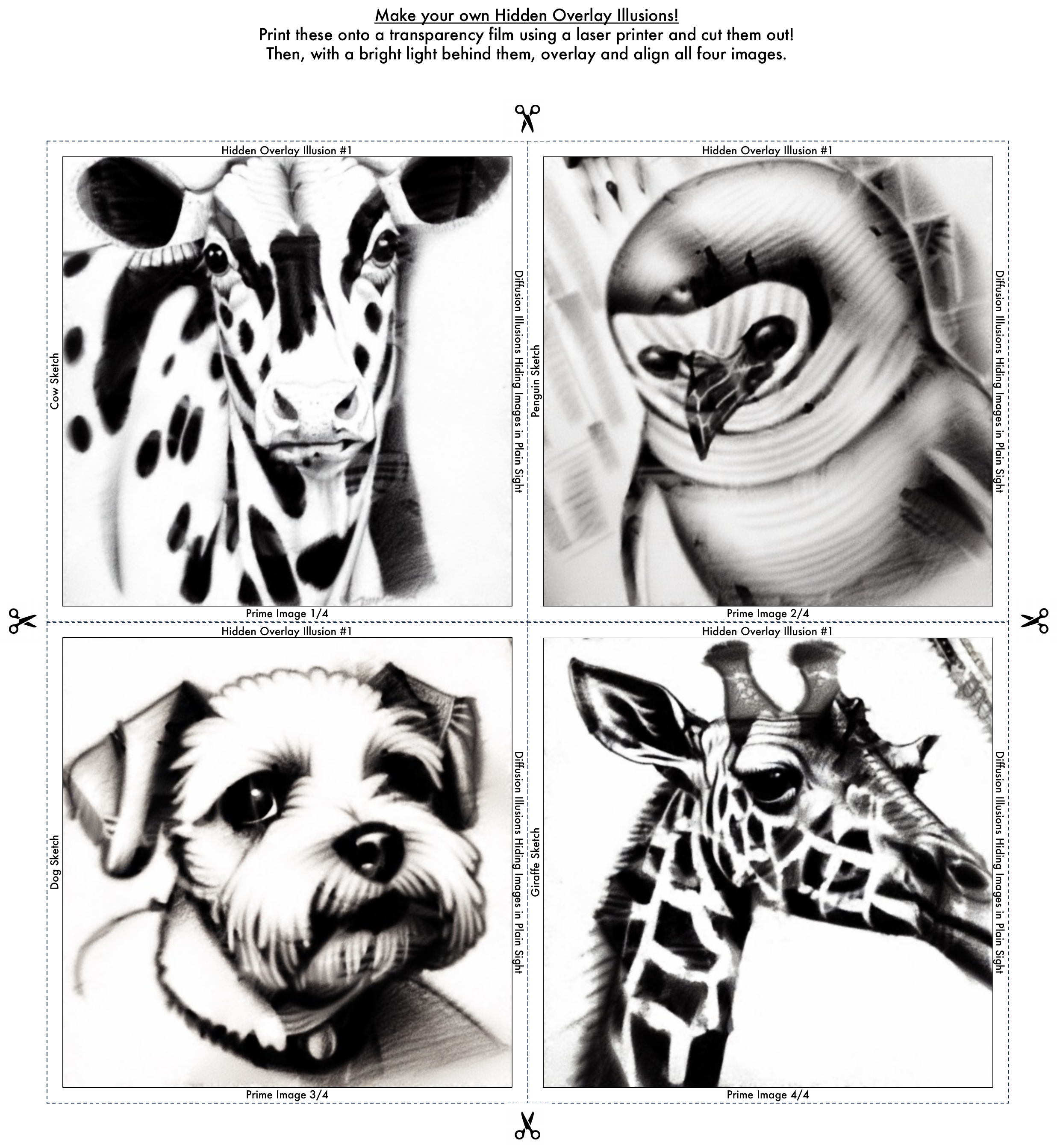}
    \label{fig:make-your-own-hidden}
\end{figure*}

\label{supp:illusion_styles}

\end{document}